\documentclass[11pt]{article}

\usepackage[preprint]{acl}

\usepackage{times}
\usepackage{latexsym}

\usepackage[T1]{fontenc}

\usepackage{microtype}

\usepackage{inconsolata}

\usepackage[most]{tcolorbox}
\usepackage{graphicx}
\usepackage{amsmath}
\usepackage{amssymb}
\usepackage{array}
\usepackage{booktabs}
\usepackage{multirow}
\usepackage{tabularx}
\usepackage{placeins}
\usepackage{tikz}
\usepackage{xurl}
\newcolumntype{Y}{>{\raggedright\arraybackslash}X}
\newcolumntype{L}[1]{>{\raggedright\arraybackslash}p{#1}}
\usetikzlibrary{arrows.meta,positioning}
\usepackage{CJKutf8}

\newcommand{\zh}[1]{%
  \begin{CJK*}{UTF8}{bsmi}#1\end{CJK*}%
}

\newcommand{\jyut}[1]{\texttt{#1}}

\title{ARIA: Audio-Driven Melody--Tone Relation Modeling for Cantonese Lyric Authoring}

\author{
Shengyu Li\thanks{These authors contributed equally to this work.}
\and
Jinting Wang\footnotemark[1]
\and
Li Liu\thanks{Corresponding author.}\\
\normalfont The Hong Kong University of Science and Technology (Guangzhou)\\
\normalfont \texttt{\{sli802,jwang644\}@connect.hkust-gz.edu.cn}\\
\normalfont \texttt{avrillliu@hkust-gz.edu.cn}
}

\usepackage{eso-pic}
\hypersetup{
  pdftitle={ARIA: Audio-Driven Melody--Tone Relation Modeling for Cantonese Lyric Authoring},
  pdfauthor={Shengyu Li; Jinting Wang; Li Liu},
  pdfsubject={Accepted for publication in Findings of EMNLP 2026; author-prepared version},
  pdfkeywords={Cantonese lyric generation, tone-melody alignment, audio-to-0243, retrieval augmentation}
}
\AddToShipoutPictureFG*{%
  \AtPageUpperLeft{%
    \put(0,-40){\makebox[\paperwidth][c]{\fontsize{8}{10}\selectfont\color{black!65}%
      Accepted for publication in Findings of EMNLP 2026. Author-prepared version.}}%
  }%
}

\begin{document}
\maketitle
\begin{abstract}

Cantonese lyric writing requires close alignment between lexical tones
and melodic pitch. Existing melody-guided lyric generation methods
typically rely on symbolic melody to generate lyrics. However, in real
songwriting scenarios, melodies are often expressed as raw singing
audio or hummed recordings, where pitch is implicit, noisy, and
unstructured, making these methods difficult to apply directly.
To address this limitation, we propose \textbf{ARIA}, a two-stage
audio-driven melody--tone relation modeling framework for Cantonese
lyric authoring that generates Cantonese lyrics from singing
recordings with provided character-level timestamps.
Specifically, we first design a
\textbf{T}ri-Stream \textbf{R}elation-\textbf{A}ware
\textbf{T}one \textbf{E}stimator (TRATE) to predict
\texttt{0243} sequences from timestamped singing audio by modeling
multi-stream acoustic cues and relational tonal structure. We then
propose a
\textbf{D}ecoupled \textbf{R}etrieval-\textbf{A}ugmented
\textbf{T}one-\textbf{C}onditioned
\textbf{L}yric \textbf{G}enerator (DRA-TCLG) to generate fluent lyrics
conditioned on predicted tonal plans with retrieval-enhanced lexical
guidance. Moreover, we construct a large-scale aligned
audio--Jyutping--\texttt{0243} dataset from real Cantonese singing
recordings to support this new task.
Experimental results demonstrate that ARIA achieves strong performance
in both \texttt{0243} prediction and tone-consistent lyric generation,
validating the effectiveness of the proposed framework.

\end{abstract}

\section{Introduction}

\label{sec:introduction}

\begin{figure}[t]
\centering
\includegraphics[width=1\linewidth]{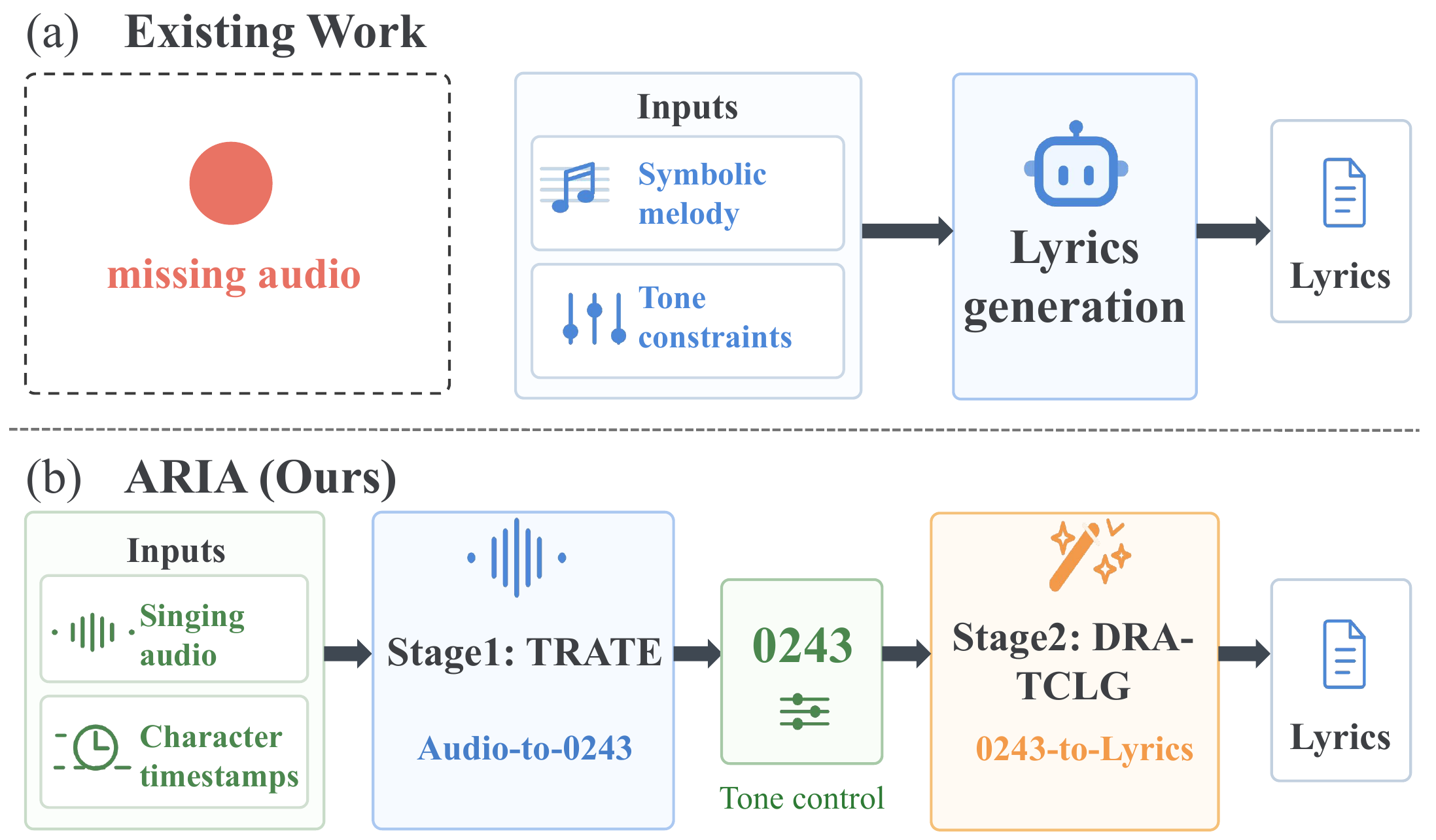}
\caption{Motivation of \textsc{ARIA}. Existing melody-aware lyric
systems typically rely on symbolic melody or manually specified tone
constraints. In contrast, \textsc{ARIA} receives singing audio
together with character-level lyric-slot boundaries, predicts an
explicit \texttt{0243} tone--melody control sequence, and uses it as
an interpretable interface for controllable Cantonese lyric
generation.}
\label{fig:motivation}
\end{figure}

Recent advances in lyric generation have substantially improved
controllability with respect to melody, rhythm, and semantic planning
\citep{sheng-etal-2021-songmass,ju-etal-2022-telemelody,
tian-etal-2023-unsupervised,qian-etal-2023-unilg,
sun-etal-2023-songrewriter,zhao-etal-2025-reffly,
yoshida-etal-2026-colyricist,ding-etal-2025-songcomposer}.
However, most existing systems assume melody is provided as clean
symbolic input (\textit{e.g.,} MIDI notes), focusing primarily on
symbolic-conditioned generation. In many practical workflows,
creators instead begin from a singing recording rather than a
structured score. We therefore study a narrower slot-aligned setting
in which the recording is accompanied by character-level lyric
boundaries.

Unlike prior work that assumes symbolic melody input or manually
specified tonal plans, we estimate the tonal plan from singing audio
while retaining character-level lyric-slot timing. Cantonese lyricists
frequently use a four-level tonal shorthand known as
\texttt{0243}\footnote{The \texttt{0243} coding scheme maps the six
Cantonese lexical tones into four canonical relative-pitch categories
following a conventional lyric-writing representation used in
Cantonese songwriting.} to bridge melodic contour and lexical tone
during lyric writing. Motivated by this practice, we introduce an
audio-driven Cantonese lyric-generation pipeline via a two-stage
framework, as illustrated in Figure~\ref{fig:motivation}(b), where
\texttt{0243} serves as an intermediate representation connecting
singing melody and lyric generation.

However, this setting introduces two key challenges. 
\textbf{First, recovering relative tonal structure from singing audio is highly non-trivial.}
Unlike conventional pitch estimation, \texttt{0243} does not represent absolute F0 values, but rather relative and line-internal tonal relationships. The same tonal category may correspond to different pitch realizations depending on melodic context. 
\textbf{Second, the mapping from \texttt{0243} sequences to lyrics is an inherently one-to-many problem.}
Even with a reliable tonal plan, multiple lyric candidates may satisfy the same \texttt{0243} pattern while differing substantially in semantics, fluency, and lexical choice.

To address these challenges, we propose \textbf{ARIA}, an audio-driven two-stage framework for Cantonese lyric authoring via melody--tone relation modeling. 
Specifically, we first design a
\textbf{T}ri-Stream \textbf{R}elation-\textbf{A}ware
\textbf{T}one \textbf{E}stimator (\textbf{TRATE}), which predicts
token-level \texttt{0243} sequences from acoustic evidence within
the supplied character slots. TRATE combines tri-stream acoustic
features with relation-aware contextual modeling, enabling reliable
recovery of relative tonal structure under realistic singing
conditions.
We further propose the \textbf{D}ecoupled \textbf{R}etrieval-\textbf{A}ugmented \textbf{T}one-\textbf{C}onditioned \textbf{L}yric \textbf{G}enerator (\textbf{DRA-TCLG}), which generates lyrics conditioned on the predicted tonal plan and retrieved lyric priors from a \texttt{0243}-lyric mapping dictionary. By incorporating retrieval-augmented lexical guidance, DRA-TCLG alleviates the one-to-many generation ambiguity and improves tonal compatibility, lexical controllability, and generation fluency.

In addition, to support this new task, we construct a large-scale
aligned audio--Jyutping--\texttt{0243} annotation resource from real
Cantonese singing recordings, providing character-level timestamps,
tone labels, and lyric alignments for training and evaluation.

\paragraph{Scope and generalizability.}
Our empirical claims are specific to Cantonese. The two-stage
audio--control--lyric decomposition, tri-stream acoustic modeling,
relation-aware sequence modeling, and delayed retrieval are
transferable at the architectural level. In contrast, the
\texttt{0243} inventory, Jyutping mapping, Cantonese tone--melody
rules, and lexical dictionary are language-specific and must be
replaced with language-appropriate controls and resources when
adapting ARIA to another language.

In summary, our contributions are threefold: 
\begin{itemize}
    \item We formulate a slot-aligned audio-driven Cantonese
    lyric-generation task that replaces externally supplied symbolic
    melody or tone-control sequences with audio-side tonal-plan
    estimation.

    \item We propose \textsc{ARIA}, a two-stage framework comprising
    TRATE for relation-aware audio-to-\texttt{0243} estimation and
    DRA-TCLG for retrieval-augmented, tone-conditioned lyric
    generation.

    \item We construct aligned Cantonese
    audio--Jyutping--\texttt{0243} annotations and evaluate the two
    stages separately and as a pipeline, including boundary
    robustness, error propagation, and human judgments.
\end{itemize}

\section{Related Work}
\begingroup
\sloppy
\label{sec:related}

\paragraph{Melody- and Tone-aware Lyric Generation.}
Prior work has explored interactive writing support, structure-aware generation, and melody-constrained lyric rewriting, including \textsc{Youling} \citep{zhang-etal-2020-youling}, \textsc{QiuNiu} \citep{zhang-etal-2022-qiuniu}, \textsc{ChipSong} \citep{liu-etal-2022-chipsong}, \textsc{UniLG} \citep{qian-etal-2023-unilg}, \textsc{SongRewriter} \citep{sun-etal-2023-songrewriter}, and \textsc{REFFLY} \citep{zhao-etal-2025-reffly}. Symbolic songwriting systems further show that explicit melody structures, such as note sequences or other symbolic melody forms, improve controllability \citep{sheng-etal-2021-songmass,ju-etal-2022-telemelody,tian-etal-2023-unsupervised,miyano-saito-2021-vocal-melody,ding-etal-2025-songcomposer}. For tonal languages, however, melody--text alignment also depends on lexical tone. Prior work has studied intermediate contour representations for Mandarin \citep{chen-teufel-2024-scansion} and tone--melody interaction in Cantonese lyric generation \citep{cheng-etal-2025-tonecraft}, while linguistic, musicological, and corpus studies suggest that sung Cantonese preserves lexical tone mainly through relative pitch relations rather than faithful spoken F0 contours \citep{wong-diehl-2002,li-choi-2016-singing-tones,yang-etal-2021-cantonese-pop,schellenberg-gick-2020}. \textsc{ARIA} is complementary to these studies but shifts the focus earlier in the pipeline: instead of assuming symbolic melody or tone-control inputs, it recovers an authoring-oriented \texttt{0243} tonal plan directly from realistic singing audio and uses it as an explicit control signal for lyric generation.

\paragraph{Audio Analysis for Lyrics and Melody.}
Lyrics-to-audio alignment and vocal melody extraction provide important tools, but they do not directly solve audio-to-\texttt{0243} mapping. Sung lyrics may not align cleanly to isolated notes, and vocal F0 estimation in polyphonic music remains difficult because of accompaniment, octave errors, vibrato, and ornamentation \citep{wong-etal-2007-lyrics-alignment,fujihara-goto-2012,salamon-etal-2014}. Large-scale Cantonese song analysis has used singing-voice extraction, ASR, character timestamps, and note extraction to quantify tone--melody relations \citep{yang-etal-2021-cantonese-pop}. \textsc{ARIA} turns this analysis-oriented problem into a controllable-generation interface: a compact \texttt{0243} sequence that can be edited, retrieved against, and consumed by a lyric generator.

\endgroup

\section{Method}
\begingroup
\sloppy
\label{sec:method}

\textsc{ARIA} decouples audio-side and text-side modeling while maintaining a unified intermediate interface, as illustrated in Figure~\ref{fig:aria-overview}. 
TRATE predicts \texttt{0243} control sequences from singing audio under known character boundaries. 
DRA-TCLG then generates lyrics conditioned on the predicted tone plan, optionally augmented with retrieved lexical candidates.
\begin{figure*}[t]
    \centering
    \includegraphics[width=0.95\textwidth]{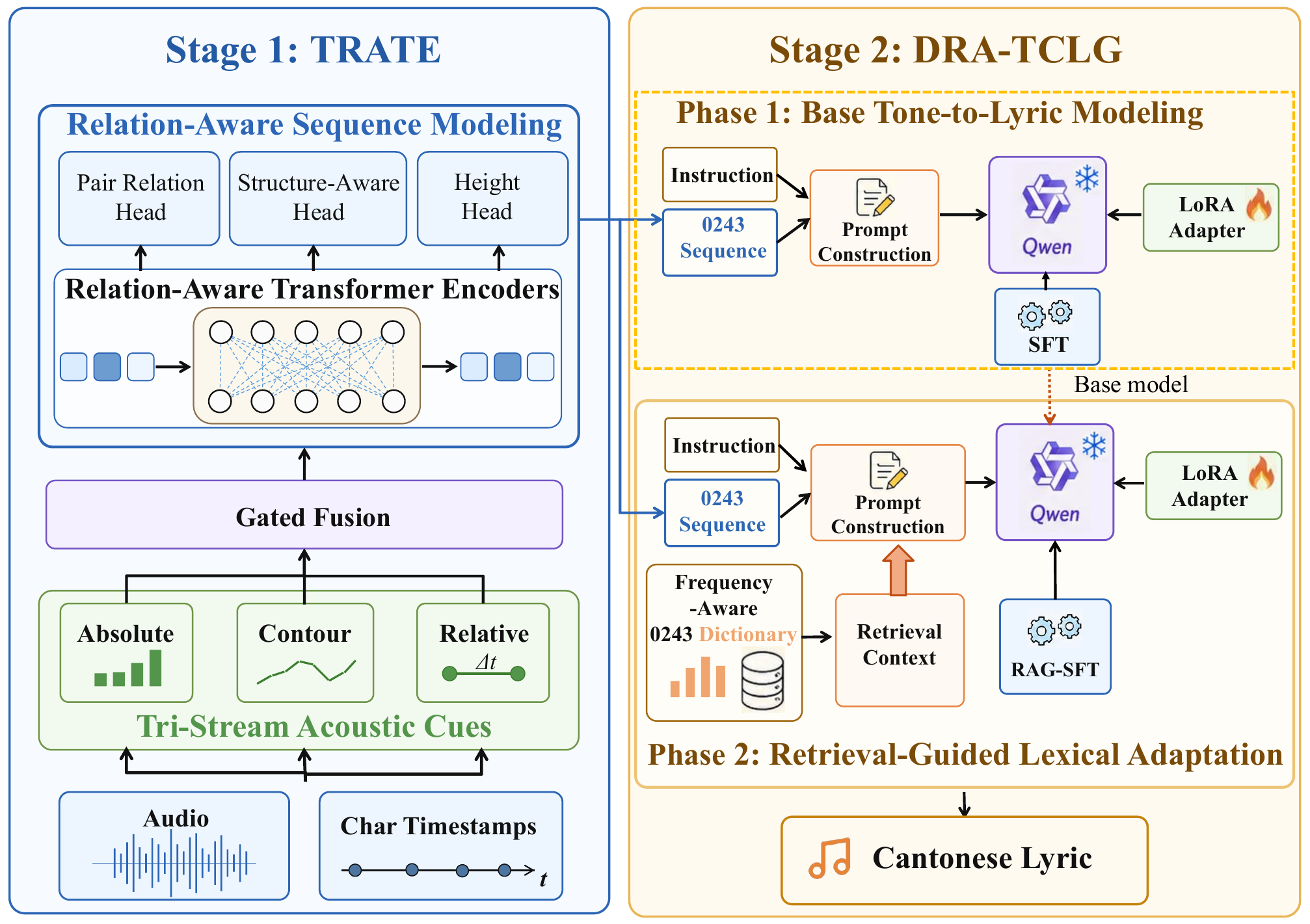}
    \caption{
    Overall architecture of \textsc{ARIA}. 
    The framework decouples audio-side tone-plan estimation and text-side lyric generation through a unified \texttt{0243} intermediate representation. 
    In Stage~1, TRATE predicts \texttt{0243} control sequences from singing audio and character-level timestamps by modeling absolute, contour, and relative acoustic streams with gated fusion and relation-aware sequence modeling. 
    In Stage~2, DRA-TCLG generates Cantonese lyrics conditioned on the predicted tonal plan, first learning a base tone-to-lyric mapping and then incorporating retrieval-enhanced lexical guidance from a frequency-aware \texttt{0243} dictionary.
    }
    \label{fig:aria-overview}
\end{figure*}

\subsection{Task Formulation}
\label{subsec:task}

Given a singing-audio segment $x$ with character-level lyric-slot
boundaries
\[
B=\{(s_i,e_i)\}_{i=1}^{N},
\]
where $N$ is the number of output character slots and $(s_i,e_i)$
denotes the start and end time of the $i$-th slot, TRATE predicts a
four-way relative-height label
\[
y_i^h \in \{1,2,3,4\},
\]
ordered from low to high, for each slot. The current system does not
discover these slots from unconstrained audio; they may be
user-provided, derived from existing annotations, or estimated by an
external lyric aligner.

We convert each height label through a fixed mapping $g$, with
$g(1)=0$, $g(2)=2$, $g(3)=4$, and $g(4)=3$, yielding
\[
P=(p_1,\ldots,p_N), \qquad p_i=g(y_i^h),
\]
where $p_i$ is the \texttt{0243} code at the $i$-th slot. For training
and evaluation, the reference \texttt{0243} sequence is derived
deterministically from the Jyutping of the aligned original lyric; at
inference time, TRATE receives neither lyric text nor Jyutping.
Because a melody may admit multiple compatible lyric realizations,
this reference denotes the tonal plan of the observed aligned lyric
rather than a unique audio-only annotation. Thus, the Stage~1 target
is a token-level authoring plan rather than frame-level F0 or a
symbolic note sequence. Appendix~\ref{app:0243-manual} describes the
label derivation.

DRA-TCLG consumes the same sequence. Given $P$ and optional controls
such as an instruction or retrieved lexical priors, it generates a
Cantonese lyric line
\[
L=(\ell_1,\ldots,\ell_N),
\]
where $\ell_i$ is the generated character aligned with $p_i$. The
line should satisfy the prescribed tone--melody tier pattern while
remaining fluent. This shared \texttt{0243} interface is compact for
generation, interpretable for human revision, and grounded in
Cantonese lyric-writing practice. Appendix~\ref{app:workflow-priors}
details how the two stages align with a real lyric-writing workflow.

\subsection{TRATE}
\label{subsec:stage1-model}

TRATE serves as the audio-side module of \textsc{ARIA}, predicting \texttt{0243} sequences from singing audio with character-level boundaries. 
As shown in the Stage~1 part of Figure~\ref{fig:aria-overview}, TRATE extracts tri-stream melodic cues from absolute, contour, and relative acoustic views, integrates them through gated fusion, and uses relation-aware Transformer encoders to capture line-internal dependencies. 
Structured prediction heads are then applied to recover the \texttt{0243} tonal control sequence.

\paragraph{Tri-Stream Feature Extraction.}
Direct F0 statistics are brittle for singing audio due to vibrato, slides, and variable token durations. 
For each character $i$, TRATE therefore builds three complementary feature vectors: absolute, contour, and relative features, denoted as $x_i^{\mathrm{abs}}$, $x_i^{\mathrm{cnt}}$, and $x_i^{\mathrm{rel}}$. 
The absolute stream summarizes token-local pitch level, stability, and duration statistics; the contour stream preserves the resampled within-token log-F0 trajectory and voiced mask; and the relative stream encodes line-internal rank, neighboring pitch gaps, local extrema, position, and relative duration. 
This separation matches the line-relative nature of \texttt{0243}, where local acoustics, within-token shape, and sentence-level ordering provide complementary evidence. 
Appendix~\ref{app:feature-details} lists the full feature set.

\paragraph{Gated Fusion.}
The three streams are encoded separately and then fused token-wise. 
Let $h_i^{\mathrm{abs}}$, $h_i^{\mathrm{cnt}}$, and $h_i^{\mathrm{rel}}$ denote the corresponding branch encodings. 
We compute the branch weights
\begin{equation}
\alpha_i =
\mathrm{softmax}\!\left(
W_g
\left[
h_i^{\mathrm{abs}}{;}\quad
h_i^{\mathrm{cnt}}{;}\quad
h_i^{\mathrm{rel}}
\right]
+b_g
\right),
\end{equation}
where $\alpha_i$ contains stream-wise gating weights for the $i$-th character, 
$[h_i^{\mathrm{abs}}{;}\quad
h_i^{\mathrm{cnt}}{;}\quad
h_i^{\mathrm{rel}}]$ denotes concatenation, and $W_g$ and $b_g$ are learnable parameters.
The fused representation is then obtained by combining the weighted branch encodings with a residual projection of their concatenation:
Let $\mathcal{M}=\{\mathrm{abs},\ \mathrm{cnt},\ \mathrm{rel}\}$.
\begin{equation}
\begin{aligned}
\tilde{h}_i
&=
\sum_{m\in\mathcal{M}}
\alpha_i^{m} h_i^{m}
+
W_r
\bigl[
h_i^{\mathrm{abs}}{;}\quad
h_i^{\mathrm{cnt}}{;}\quad
h_i^{\mathrm{rel}}
\bigr].
\end{aligned}
\end{equation}
where $\alpha_i^{m}$ is the weight for stream $m$ and $W_r$ is a learnable residual projection. 
This lets the model adapt to token-level reliability: a sustained syllable may rely on contour shape, whereas a short token may rely more on rank and local statistics.

\paragraph{Relation-Aware Sequence Modeling.} The fused sequence $\tilde{H}=\{\tilde{h}_1,\ldots,\tilde{h}_N\}$ is processed by a Transformer encoder \citep{vaswani2017attention}. Since \texttt{0243} is line-internal and ordinal, we add pairwise melodic relation bias to self-attention: 
\begin{equation} 
e_{ij}^{(l,a)} =
\left(\left(q_i^{(l,a)}\right)^\top k_j^{(l,a)}\right) \big/ \sqrt{d}
+ u_{l,a}^{\top}\phi(r_{ij}).
\end{equation}
where $e_{ij}^{(l,a)}$ is the attention logit from token $i$ to token $j$ in layer $l$ and head $a$, $q_i^{(l,a)}$ and $k_j^{(l,a)}$ are the query and key vectors, and $d$ is the attention-head dimension. The vector $r_{ij}$ summarizes median log-F0 gap, percent-rank gap, token distance, sign, and relative position between tokens $i$ and $j$; $\phi(\cdot)$ projects this relation vector, and $u_{l,a}$ is a learnable relation-bias vector for layer $l$ and head $a$. On top of the contextualized representation $z_i$ produced by the Transformer encoder, TRATE predicts the main height label together with contour-family, register, and pairwise relation heads. The contour and register heads factor Cantonese tonal supervision into structured components, while the pairwise head predicts short-range ordering relations over nearby tokens. Rather than treating the six-category Cantonese tone label as a free flat tone-6 target, we derive a height distribution from the structured tonal pathway and regularize it toward the main level-4 prediction. Training combines the main height loss, auxiliary classification losses, a height-consistency regularizer, and pitch-shift transposition consistency.

\subsection{DRA-TCLG}
\label{subsec:stage2-model}

DRA-TCLG serves as the text-side module of \textsc{ARIA}, converting the \texttt{0243} sequence predicted by TRATE into Cantonese lyrics. 
Although large language models can generate fluent text, they do not explicitly encode the correspondence between \texttt{0243} patterns and Cantonese lexical-tone choices. 
We therefore formulate DRA-TCLG as a tone-conditioned generator and introduce compact dictionary priors only after learning a stable tone-to-lyric mapping. 
Directly applying retrieval-augmented generation (RAG) from the beginning can introduce noisy candidate sets that destabilize early-stage training and weaken tone--lyric alignment. 
To address this issue, we adopt a decoupled two-phase training strategy, as illustrated in the Stage~2 part of Figure~\ref{fig:aria-overview}.

\paragraph{Phase 1: Base Tone-to-Lyric Modeling.}
We start from Qwen3.5-4B~\citep{qwen3.5} and train a causal language model to map the \texttt{0243} control sequence $P$ to a target lyric line $y=(y_1,\ldots,y_T)$ without retrieval, where $T$ is the number of lyric tokens:
\begin{equation}
\begin{aligned}
p_{\text{base}} &= [\text{instruction};P],\\
\mathcal{L}_{\text{base}}
&= -\sum_{t=1}^{T}\log p_{\theta}(y_t \mid y_{<t}, p_{\text{base}}).
\end{aligned}
\end{equation}
Here, $p_{\text{base}}$ is the Phase~1 prompt, $y_t$ is the target token at step $t$, $y_{<t}$ denotes previous target tokens, and $\theta$ denotes the model parameters.

Only lyric tokens are supervised during training. 
This phase encourages the model to first acquire a direct mapping among the \texttt{0243} sequence length, rhythmic structure, and Cantonese lexical realization, without being distracted by noisy or incomplete candidate lists. 
It also provides the second phase with a stable initialization, so retrieval is introduced as lexical refinement and constraint enhancement rather than being responsible for learning the tone-to-lyric mapping from scratch.

\paragraph{Phase 2: Retrieval-Guided Adaptation.}
The second phase adds lexical priors from 0243.hk. 
We first build a frequency-aware \texttt{0243}-to-lyrics dictionary by counting candidate occurrences in the Fei Tsui training lyrics and reordering each code entry by corpus frequency. 
For an input \texttt{0243} sequence $P$, we split it into code blocks, enumerate sub-codes with sliding windows of length 2--4, and retrieve candidate words from this dictionary. 
We prioritize 2- and 3-character candidates because they are common in Cantonese lyrics, retain only sparse 4-character candidates to avoid over-constraining generation, and discard single-character candidates because they are dominated by frequent function words. 
The retrieved context is serialized as aligned code--candidate pairs:
\begin{equation}
\begin{aligned}
R &= \{(c_i,\mathcal{C}(c_i))\}_{i=1}^{M},\\
r &= \operatorname{serialize}(R),\\
p_{\text{rag}} &= [\text{instruction};r;P],
\end{aligned}
\end{equation}
where $c_i$ is the $i$-th retrieved sub-code,
$\mathcal{C}(c_i)$ is its candidate-word set, $M$ is the number of
retrieved sub-codes, $R$ is the structured retrieval result, $r$ is
its serialized prompt form, and $p_{\text{rag}}$ is the
retrieval-augmented prompt.

Starting from the Phase~1 model, we fine-tune LoRA adapters~\citep{hu2022lora} with the backbone frozen:
\begin{equation}
\mathcal{L}_{\text{rag}}
=-\sum_{t=1}^{T}\log p_{\theta'}(y_t\mid y_{<t},p_{\text{rag}}),
\end{equation}
where $\theta'$ denotes the adapted model parameters, and $y_t$, $y_{<t}$, and $T$ follow the same definitions as in Phase~1.
This decoupling prevents early optimization from overfitting to noisy retrieval candidates while still letting the final model use domain-specific lexical priors. 
The retrieval module is designed for prompt density rather than maximum recall: it gives DRA-TCLG a compact set of plausible lexical anchors while leaving the language model free to ignore candidates that do not fit the semantic context. 
Appendix~\ref{app:DRA-TCLG-retrieval} gives the exact prompt format and retrieval rules.

\endgroup

\section{Experiments}
\label{sec:experiments}

\subsection{Dataset}
\label{subsec:dataset}

Following ToneCraft's Fei Tsui setting \citep{cheng-etal-2025-tonecraft}, we collect Cantonese lyrics from Fei Tsui\footnote{\url{https://www.feitsui.com/}}, which contains over 6k songs. 
\textsc{ARIA} uses three sources: lyrics and matched singing recordings for Stage~1 audio-to-\texttt{0243} learning, and lyrics plus 0243.hk\footnote{\url{https://www.0243.hk/}} lexical resources for Stage~2 \texttt{0243}-to-lyric generation. 
After filtering, Stage~1 contains over 100K aligned lyric--audio line samples from over 3.4K recordings, split by song into train/dev/test sets with an approximate 0.8/0.1/0.1 ratio to avoid song-level leakage. 
Stage~2 contains over 170K Fei Tsui \texttt{0243}-to-lyric pairs, split by 0.9/0.05/0.05. 
For comparing oracle \texttt{0243}-to-lyrics generation with the full audio-to-lyrics pipeline, we build a shared 500-sample test subset from the intersection of the Stage~1 and Stage~2 test sets, matched by normalized \texttt{0243} sequences. 
We also retrieve over 140K unique 2--4 character candidates from 0243.hk. 
Due to copyright restrictions, we do not redistribute the original recordings; instead, we release non-audio Stage~1 annotations, including source identifiers, splits, character boundaries, \texttt{0243} and level-4 labels, Jyutping, pitch metadata, and preprocessing scripts. 
Exact statistics, construction details, and the released format are provided in Appendices~\ref{app:dataset-stats} and~\ref{sec:released-annotation-format}.

\subsection{Experimental Setup}
\label{subsec:exp-setup}

\paragraph{TRATE.}
It predicts token-level \texttt{0243} from singing audio and character
timestamps. We compare it with two quartile rules and two Qwen3.5
baselines~\citep{qwen3.5}, and report TokAcc, Macro-F1, Within
$\pm1$, and normalized edit distance (NED). These metrics cover exact
labels, class balance, musically adjacent near misses, and downstream
sequence stability; feature and metric details are in
Appendices~\ref{app:feature-details} and~\ref{app:withinpmone}.
We optimize with AdamW~\citep{loshchilov2019decoupled}, using linear
warmup followed by cosine learning-rate decay. Only the main height
head uses label smoothing, while EMA is enabled. The TRATE result in
Table~\ref{tab:main-results} is the arithmetic mean over three
independent training seeds (13, 42, and 3407).
Appendix~\ref{app:stage1-extra} gives the full objective.

\paragraph{DRA-TCLG.}
It uses frozen Qwen3.5-4B with LoRA adapters \citep{qwen3.5,hu2022lora}; it first learns a base \texttt{0243}-to-lyric mapping and then receives retrieval-guided adaptation from 0243.hk candidates. We compare it with in-domain ToneCraft-Qwen2-7B, SmBART, and SongNet variants under the same Fei Tsui distribution \citep{cheng-etal-2025-tonecraft,chen-teufel-2024-scansion,li-etal-2020-rigid}. Following ToneCraft \citep{cheng-etal-2025-tonecraft}, we evaluate alignment with Harmony and Consistency, and diversity with embedding similarity and entropy-based MaD/MiD metrics; training settings and baseline adaptations are in Appendix~\ref{app:exp-details}.

This setup separates two evaluation regimes. The \texttt{0243}-to-lyrics generation results measure the generator under ideal control inputs, while the audio-to-lyrics subset measures the full \textsc{ARIA} pipeline when \texttt{0243} must first be recovered from singing audio. Keeping both regimes visible lets us distinguish generator quality from error propagation through the Stage~1 interface.

All \texttt{0243}-to-lyrics systems in
Tables~\ref{tab:main-results-lyrics}
and~\ref{tab:main_results_lyrics_500}, including
ToneCraft-Qwen2, receive reference \texttt{0243} sequences derived
from the original lyrics. Only TRATE+DRA-TCLG is evaluated from
singing audio with character-level timestamps, so its result includes
the Stage~1 audio-to-\texttt{0243} prediction process. 
The approximate main-training cost is 2.8 hours for TRATE on one
NVIDIA A800-SXM4-80GB GPU and 7 hours for DRA-TCLG on one NVIDIA H20
GPU. These figures exclude baseline training, ablations, evaluation,
and inference.

\subsection{Main Results}
\label{subsec:main-results}

\begin{table}[t]
\centering
\footnotesize
\setlength{\tabcolsep}{3.0pt}
\renewcommand{\arraystretch}{1.05}
\begin{tabular*}{\linewidth}{@{\extracolsep{\fill}}lccccc@{}}
\toprule
Model 
& Acc $\uparrow$ 
& F1 $\uparrow$ 
& $\pm1$ $\uparrow$ 
& NED $\downarrow$ 
& Params \\
\midrule
Rule RQ 
& 0.373 
& 0.360 
& 0.749 
& 0.583 
& 0 \\
Rule AQ 
& 0.370 
& 0.357 
& 0.747 
& 0.585 
& 0 \\
Qwen Rel. 
& 0.501 
& 0.471 
& 0.835 
& 0.460 
& \shortstack{4.57B /\\ 32.46M tr.} \\
Qwen 3-Br. 
& \underline{0.565} 
& \underline{0.537} 
& \underline{0.844} 
& \underline{0.398} 
& \shortstack{4.57B /\\ 32.46M tr.} \\
TRATE 
& \textbf{0.644} 
& \textbf{0.615} 
& \textbf{0.885} 
& \textbf{0.347} 
& 2.37M \\
\bottomrule
\end{tabular*}
\caption{Audio-to-\texttt{0243} results. Acc, F1, and $\pm1$
denote TokAcc, Macro-F1, and Within $\pm1$, respectively. Higher is
better except for NED. The TRATE row reports the mean over three
training seeds (13, 42, and 3407). Rule-based methods require no
trainable parameters. For Qwen baselines, ``tr.'' denotes trainable
parameters. The best results are highlighted in \textbf{bold}, and
the second-best results are \underline{underlined}.}
\label{tab:main-results}
\end{table}

\noindent\textbf{Audio-to-\texttt{0243}.}
Table~\ref{tab:main-results} shows that TRATE is best on all four metrics, reaching 0.644 TokAcc, 0.615 Macro-F1, 0.885 Within $\pm1$, and 0.347 NED. Compared with the strongest Qwen3.5 baseline, it improves TokAcc by 0.079 and reduces NED by 0.051 while using a 2.37M-parameter specialized encoder. The rule baselines remain around 0.37 TokAcc, showing that raw quartile splitting is too coarse for sung Cantonese, while the gap between Qwen Rel. and Qwen 3-Br. suggests that absolute, contour, and relative cues are complementary. This matters for the full system because DRA-TCLG consumes a complete line-level \texttt{0243} plan rather than isolated token labels; a lower edit distance therefore makes the downstream control signal more stable. More detailed error analysis is in Appendix~\ref{app:main-result-details}.

\begin{table*}[t] 
\centering
\footnotesize
\setlength{\tabcolsep}{2.4pt}
\renewcommand{\arraystretch}{1.02}
\begin{tabular*}{\textwidth}{@{\extracolsep{\fill}}lcccccccc@{}}
\toprule
\multirow{2}{*}{Method}
& \multicolumn{2}{c}{Alignment}
& \multicolumn{6}{c}{Diversity} \\
\cmidrule(lr){2-3}
\cmidrule(lr){4-9}
& Harmony $\uparrow$
& Consistency $\uparrow$
& Avg Sim $\downarrow$
& Min Sim $\downarrow$
& MaD1 $\uparrow$
& MaD2 $\uparrow$
& MiD1 $\uparrow$
& MiD2 $\uparrow$ \\
\midrule
SongNet & 0.4846 & 0.0375 & \underline{0.4911} & 0.0479 & 0.9354 & \underline{0.9862} & \underline{0.0253} & \underline{0.3341} \\
SmBART & 0.8186 & 0.6755 & \textbf{0.4277} & \textbf{0.0008} & 0.8830 & 0.9410 & \textbf{0.0297} & \textbf{0.4258} \\
ToneCraft & \underline{0.9684} & \underline{0.9347} & 0.5036 & 0.0541 & \textbf{0.9552} & 0.9855 & 0.0134 & 0.1403 \\
DRA-TCLG (Ours) & \textbf{0.9740} & \textbf{0.9421} & 0.5053 & \underline{0.0193} & \underline{0.9544} & \textbf{0.9944} & 0.0190 & 0.2967 \\
\bottomrule
\end{tabular*}
\caption{\texttt{0243}-to-lyric generation comparison across alignment and diversity metrics.}
\label{tab:main-results-lyrics}
\end{table*}

\begin{table*}[t]
\centering
\footnotesize
\setlength{\tabcolsep}{2.4pt}
\renewcommand{\arraystretch}{1.02}

\begin{tabular*}{\textwidth}{@{\extracolsep{\fill}}lcccccccc@{}}
\toprule
\multirow{2}{*}{Method} 
& \multicolumn{2}{c}{Alignment} 
& \multicolumn{6}{c}{Diversity} \\
\cmidrule(lr){2-3}
\cmidrule(lr){4-9}
& Harmony $\uparrow$
& Consistency $\uparrow$
& Avg Sim $\downarrow$
& Min Sim $\downarrow$
& MaD1 $\uparrow$
& MaD2 $\uparrow$
& MiD1 $\uparrow$
& MiD2 $\uparrow$ \\
\midrule
\multicolumn{9}{l}{
\textit{\texttt{0243}-to-Lyrics
(reference \texttt{0243} input)}} \\
SongNet 
& 0.5000 
& 0.0353 
& \underline{0.4699} 
& 0.1034 
& 0.9487 
& 0.9894 
& \underline{0.1896} 
& \underline{0.7352} \\
SmBART 
& 0.8277 
& 0.6373 
& \textbf{0.4138} 
& \textbf{0.0877} 
& 0.9037 
& 0.9524 
& \textbf{0.2273} 
& \textbf{0.7849} \\
ToneCraft 
& 0.9747
& 0.9466
& 0.4957 
& 0.0997 
& 0.9612 
& 0.9882 
& 0.0945 
& 0.3763 \\
DRA-TCLG (Ours)
& \textbf{0.9841}
& \textbf{0.9687} 
& 0.4849
& 0.1100
& \underline{0.9630}
& \underline{0.9951}
& 0.1345
& 0.6246\\
\midrule
\multicolumn{9}{l}{
\textit{Audio-to-Lyrics
(singing audio + character timestamps)}} \\
TRATE+DRA-TCLG (Ours)
& \underline{0.9786}
& \underline{0.9528}
& 0.4929
& \underline{0.0977}
& \textbf{0.9790}
& \textbf{0.9980}
& 0.1359
& 0.6545\\
\bottomrule
\end{tabular*}
\caption{Comparison on the shared 500-sample test subset under two
input regimes. The \texttt{0243}-to-lyrics systems receive reference
\texttt{0243} sequences derived from the original lyrics, whereas
TRATE+DRA-TCLG receives singing audio with character-level timestamps
and first predicts its own \texttt{0243} plan. Harmony and Consistency
are computed against the control sequence supplied to each generator.
The table therefore provides a system-level cross-regime comparison
rather than a same-input comparison.}
\label{tab:main_results_lyrics_500}
\end{table*}

\noindent\textbf{\texttt{0243}-to-Lyrics.}
Table~\ref{tab:main-results-lyrics} shows that DRA-TCLG achieves the strongest tone--melody alignment, with a Harmony score of 0.9740 and a Consistency score of 0.9421. 
Compared with the strong ToneCraft-Qwen2 baseline, DRA-TCLG further improves both metrics, showing that retrieval-guided tone-conditioned modeling adds benefits beyond direct tone-aware generation. 
SmBART obtains competitive diversity but lower alignment, suggesting that lexical variety and tone--melody compatibility can be partially competing objectives in Cantonese lyric generation. 
Overall, DRA-TCLG achieves a better trade-off by closely following the \texttt{0243} plan while preserving strong lexical diversity. 
Appendix~\ref{app:main-result-details} provides additional analysis, and Appendix~\ref{app:crosslingual-stress} reports the Mandarin-to-Cantonese stress test.
The comparison also shows that this task is more than standard Chinese lyric generation with an added prompt. 
SongNet and SmBART produce fluent and diverse lyrics, but do not reliably follow the prescribed \texttt{0243} structure. 
Although ToneCraft-Qwen2 reaches strong alignment through explicit tone-aware modeling, DRA-TCLG further improves Harmony and Consistency by retrieving tone-compatible candidates as lexical priors. 
This suggests that a shared \texttt{0243} control interface with retrieval-guided adaptation provides stronger control than direct tone-aware generation or general lyric-generation ability alone.

\noindent\textbf{Audio-to-Lyrics.}
Table~\ref{tab:main_results_lyrics_500} intentionally compares the
complete \textsc{ARIA} pipeline with \texttt{0243}-conditioned systems
under two different input regimes. SongNet, SmBART,
ToneCraft-Qwen2, and oracle DRA-TCLG receive reference
\texttt{0243} sequences derived from the original lyrics.
TRATE+DRA-TCLG instead receives singing audio with character-level
timestamps and must first infer the \texttt{0243} plan. This comparison
therefore evaluates system-level performance when a reference tone plan
is unavailable at inference time, rather than constituting a
same-input comparison.

Under this harder input regime, the complete pipeline achieves
0.9786 Harmony and 0.9528 Consistency while obtaining the best
MaD1/MaD2 scores. These alignment values show that DRA-TCLG remains
highly faithful to the plan inferred by TRATE; they do not by
themselves measure agreement with the lyric-derived reference plan.
We therefore analyze reference-level Stage~1 error propagation below.

\paragraph{Stage~1 error propagation.}
In a separate paired-generation intervention on the same 500-item set, we regenerate one output under the reference plan and one under
the Stage~1-predicted plan. Because this intervention uses a separate
matched generation run, its reference-input aggregate is reported
independently of the oracle result in
Table~\ref{tab:main_results_lyrics_500}. Replacing the reference
\texttt{0243} input with Stage~1 predictions lowers
final-vs-reference Harmony from 0.9742 to 0.8482
($\Delta=-0.1260$). 
Generation uses stochastic decoding, so the independently regenerated
reference-input aggregate need not be numerically identical to the
oracle aggregate in Table~\ref{tab:main_results_lyrics_500}.
The reference-input and predicted-input conditions produce identical extracted output
\texttt{0243} sequences in only 15.2\% of cases, and Stage~1 NED
strongly correlates with final-output \texttt{0243} NED
(Spearman's $\rho=0.8520$).

The loss is concentrated in the structure of Stage~1 errors.
Adjacent-only errors retain Harmony/Consistency of
0.8404/0.7810 ($n=265$), whereas predictions containing a
non-adjacent error fall to 0.7568/0.5592 ($n=131$).
Harmony further decreases from 0.8507 for isolated errors to
0.7619 for maximum error runs of length two and 0.6843 for runs
of length at least three. By contrast, final-vs-predicted-plan
Harmony remains 0.9716--0.9738 across the adjacent-only and
non-adjacent groups. Stage~2 therefore generally realizes rather
than repairs the supplied plan.

\paragraph{Qualitative cases.}
For an isolated adjacent substitution, the lyric-derived reference,
TRATE-predicted, and generated-output plans are
\texttt{0024433334424}, \texttt{0224433334424}, and
\texttt{0224433334424}; final-vs-reference Harmony/Consistency remains
0.9514/0.9909. In a severe eight-position run,
\texttt{00240024} is predicted and realized as \texttt{44334433},
reducing final-vs-reference Harmony to 0.1935. The corresponding reference and generated lyrics, Jyutping transliterations, English glosses, and plan-level scores are provided
in Appendix~\ref{app:error-cases}.

\subsection{Ablation Study}
\label{subsec:ablation}

\begin{figure}[t]
\centering
\includegraphics[width=1\linewidth]{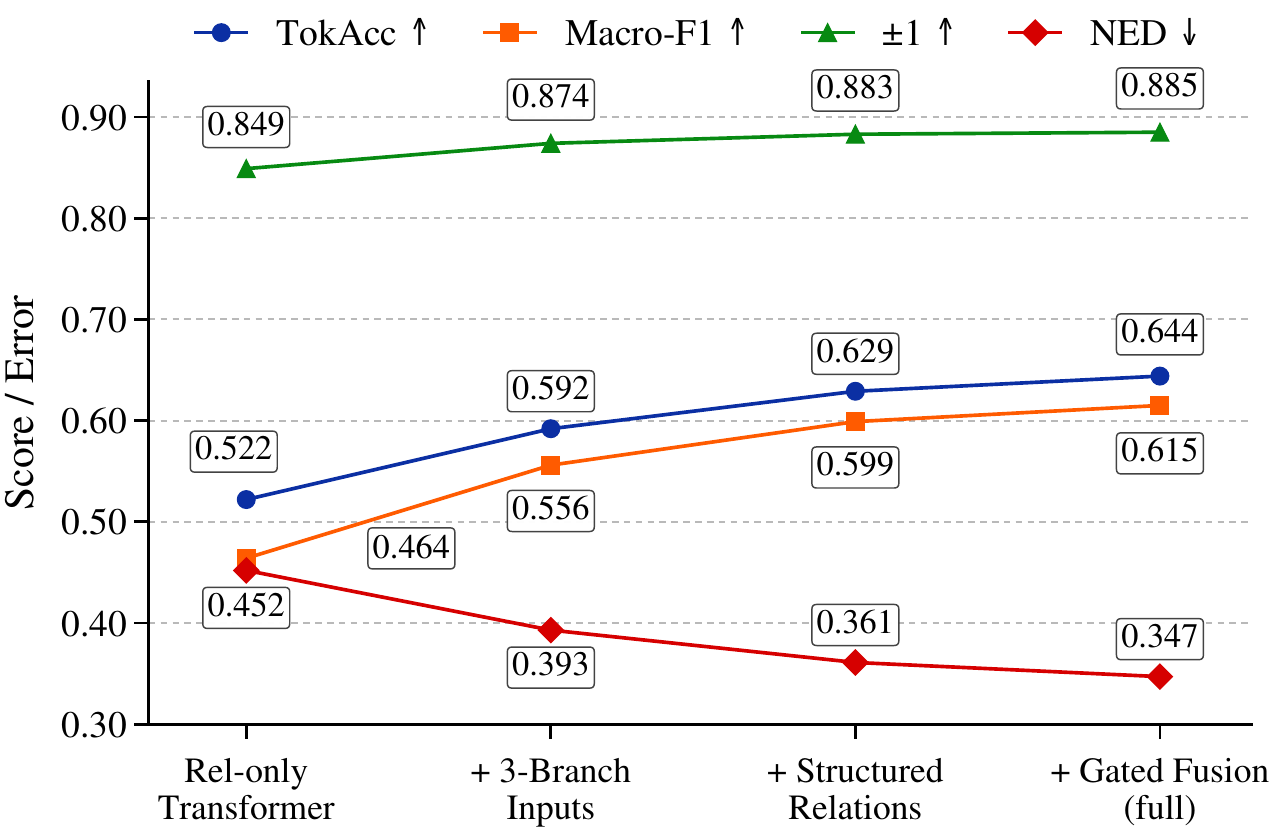}
\caption{TRATE ablation. The progression isolates richer token features, structured relation-aware modeling, and token-wise gated fusion.}
\label{fig:stage1-ablation}
\end{figure}
\noindent\textbf{Audio-to-\texttt{0243}.} Figure~\ref{fig:stage1-ablation} shows a monotonic gain across the TRATE stack. The three-branch representation gives the largest jump over the rel-only Transformer, improving TokAcc by 0.070 and reducing NED by 0.059. Structured relations further improve both exact and sequence-level metrics, supporting the claim that \texttt{0243} recovery is relational rather than purely local. Gated fusion gives the best final scores, which indicates that the most reliable evidence source varies across tokens. These results support the design choice of separating absolute, contour, and relative cues before relation-aware sequence modeling; detailed analysis is in Appendix~\ref{app:ablation-details}.

The component trend also matches the acoustic difficulty of sung Cantonese. A sustained note, a gliding syllable, and a neighboring pair with similar median height can require different evidence. Keeping the streams separate before fusion lets the model use local pitch level, within-token contour, and line-internal ordering as complementary signals instead of forcing all evidence into one pooled pitch feature.

\paragraph{Boundary-noise Robustness.}
Because TRATE relies on character-level timestamps, we additionally
evaluate its sensitivity to boundary noise by perturbing character
start and end times before feature extraction. Unlike the three-seed
average in Table~\ref{tab:main-results}, this analysis fixes the best
EMA checkpoint from training seed 3407. Each nonzero perturbation
condition is averaged over three independently sampled perturbation
seeds, while the clean condition is evaluated once through the same
feature-recomputation pipeline.

Performance remains stable under $\pm25$ ms perturbation and degrades
moderately under larger offsets. At $\pm50$ ms, Acc and F1 decrease
from 0.6459 and 0.6173 to 0.6391 and 0.6102, respectively, while
Within $\pm1$ changes only from 0.8840 to 0.8837. Full results and
protocol details are provided in
Appendix~\ref{app:boundary-robustness}.

\begin{table}[t]
\centering
\footnotesize
\setlength{\tabcolsep}{3.6pt}
\renewcommand{\arraystretch}{1.05}
\begin{tabular*}{\linewidth}{@{\extracolsep{\fill}}lccc@{}}
\toprule
\multirow{2}{*}{Metric}
& \multicolumn{3}{c}{Method} \\
\cmidrule(lr){2-4}
& Base & One-Stage & Two-Stage \\
\midrule
\multicolumn{4}{@{}l}{\textit{Alignment}} \\
Harmony $\uparrow$
& \underline{0.9104}
& 0.6449
& \textbf{0.9740} \\
Consistency $\uparrow$
& \underline{0.7619}
& 0.1579
& \textbf{0.9421} \\
\midrule
\multicolumn{4}{@{}l}{\textit{Diversity}} \\
Avg Sim $\downarrow$
& 0.5269
& \textbf{0.4912}
& \underline{0.5053} \\
Min Sim $\downarrow$
& 0.0731
& \underline{0.0257}
& \textbf{0.0193} \\
MaD1 $\uparrow$
& 0.9494
& \textbf{0.9548}
& \underline{0.9544} \\
MaD2 $\uparrow$
& \underline{0.9888}
& \textbf{0.9944}
& \textbf{0.9944} \\
MiD1 $\uparrow$
& 0.0130
& \textbf{0.0193}
& \underline{0.0190} \\
MiD2 $\uparrow$
& 0.1368
& \underline{0.2670}
& \textbf{0.2967} \\
\bottomrule
\end{tabular*}
\caption{Ablation study of DRA-TCLG.}
\label{tab:ablation-rag}
\end{table}

\noindent\textbf{\texttt{0243}-to-Lyrics.}
Table~\ref{tab:ablation-rag} shows that retrieval helps only after the
base tone-to-lyric mapping is stable. One-Stage RAG increases some
diversity scores but sharply degrades Harmony and Consistency, whereas
Two-Stage RAG achieves the best alignment while retaining competitive
diversity. The contrast suggests that raw candidate lists are useful
lexical priors but harmful early supervision: before the model has
learned the \texttt{0243}-to-lyric relation, retrieval noise can
distract it from tone alignment. The decoupled strategy avoids this by
first learning the base mapping, then using 0243.hk candidates as
optional lexical refinement. On the training set, at least one retrieved candidate appears
verbatim in approximately 37.7\% of target lyrics. This limited exact
hit rate indicates that retrieval acts as a sparse lexical prior
rather than an answer bank. Two representative cases further
illustrate this behavior: for input \texttt{2433334}, the model uses
two retrieved candidates and still matches the \texttt{0243} plan
exactly; for input \texttt{22342433}, it uses none of the retrieved
candidates and also matches the plan exactly. Retrieved candidates
are therefore optional lexical suggestions rather than copied
answers.
Appendix~\ref{app:retrieval-cases} gives representative use and
non-use cases, and Appendix~\ref{app:ablation-details} gives the
fuller trade-off analysis.

This behavior is important for the proposed workflow. The retrieval module should expand the writer's lexical search space, not override the tone plan. Two-Stage RAG is therefore consistent with the intended authoring process: the model first learns how a line-length and code pattern constrain a lyric line, then uses frequent tone-compatible candidates when they support fluent realization.

The two ablations are complementary. TRATE improves the stability of the sequence-level interface, and DRA-TCLG shows that lexical priors are most useful after the interface has been learned. This is why the evaluation reports both audio-side sequence metrics and text-side alignment metrics: the system's performance depends on whether the recovered 0243 plan remains usable for retrieval-guided generation.

\subsection{User Study}
\label{sec:user-study}

We conduct a human evaluation on 40 non-overlapping items from the
shared 500-item test subset, following the evaluation
dimensions of ToneCraft~\citep{cheng-etal-2025-tonecraft}.
Each item is evaluated under the same blind protocol by 12
Cantonese-speaking raters. All four systems are conditioned on the same reference
\texttt{0243} sequence; the vocal recording is presented only to
raters as a melody reference. 

As shown in Table~\ref{tab:user-study-main}, DRA-TCLG obtains the
highest descriptive mean on all five dimensions and receives the most
valid preference votes. We report these results descriptively and do
not claim statistical significance. Full study details are provided
in Appendix~\ref{app:user-study}.

\begin{table}[t]
\centering
\footnotesize
\setlength{\tabcolsep}{0pt}
\begin{tabular*}{\columnwidth}
{@{\extracolsep{\fill}}lcccc@{}}
\toprule
Metric & DRA-TCLG & ToneCraft & SmBART & SongNet \\
\midrule
Tone--Mel. & \textbf{3.185} & \underline{2.967} & 2.654 & 2.029 \\
Rhythm     & \textbf{3.256} & \underline{2.998} & 2.671 & 2.152 \\
Canto.     & \textbf{3.233} & \underline{3.167} & 2.710 & 2.077 \\
Lyric      & \textbf{3.279} & \underline{3.054} & 2.756 & 2.202 \\
Sing.      & \textbf{3.167} & \underline{3.015} & 2.627 & 2.027 \\
Pref.      & \textbf{160} & \underline{142} & 89 & 22 \\
\bottomrule
\end{tabular*}
\caption{Human evaluation on 40 items with 12 Cantonese-speaking
raters. Values are descriptive mean ratings except Pref., which
reports 413 valid preference votes; 56 uncertain and 11 blank
responses are excluded. Higher is better.}
\label{tab:user-study-main}
\end{table}

\section{Conclusion}
\label{sec:conclusion}

We introduced \textsc{ARIA}, a slot-aligned audio-driven framework
for Cantonese lyric authoring that connects singing audio, the
traditional \texttt{0243} representation, TRATE, and DRA-TCLG. TRATE
shows that token-level audio-to-\texttt{0243} mapping is difficult but
learnable when acoustic, contour, and line-internal relations are
modeled together, while DRA-TCLG uses the resulting plan for
tone-conditioned generation with compact lexical priors. Error
analysis shows that Stage~2 usually realizes rather than repairs the
supplied plan, making boundary quality and Stage~1 prediction the main
bottlenecks. The architecture provides a human-readable interface
between audio analysis and controllable lyric writing, while the
present empirical claims and resources remain Cantonese-specific.

\section*{Limitations}

Our current formulation assumes character-level lyric-slot boundaries
for singing audio. These slots may be user-specified, derived from
existing annotations, or estimated by an external lyric aligner.
Thus, \textsc{ARIA} removes the need for an externally supplied
symbolic melody or tonal plan, but it does not discover slot
boundaries from unconstrained audio. Because we did not measure
preparation time, we do not claim a universal cost advantage over
symbolic-input systems. Reference \texttt{0243} labels are
deterministically derived from the Jyutping of aligned original
lyrics; they therefore represent the tonal plans of observed lyric
realizations rather than unique audio-only annotations.

Our empirical claims are Cantonese-specific. The audio--control--text
decomposition, tri-stream acoustic modeling, relation-aware sequence
modeling, and delayed retrieval are architecture-level components.
The \texttt{0243} inventory, Jyutping-based supervision, Cantonese
tone--melody rules, and lexical resources are language-specific and
must be redesigned for another language. Stage~2 generally realizes
rather than repairs an incorrect Stage~1 plan, so clustered and
non-adjacent prediction errors remain consequential. Finally, the human evaluation remains modest in scale. Although the
main TRATE result is averaged over three training seeds, we do not
report confidence intervals or complete seed-level variance estimates
for every model and analysis.

\section*{Ethical Considerations}
Original singing recordings are copyrighted commercial music files and are therefore excluded from public release. 
We release only non-audio annotations, including source identifiers, character-level timestamps, Jyutping-derived tonal labels, \texttt{0243} labels, split files, derived pitch metadata, and reconstruction scripts. 
This design supports reproducibility for users with legal access to the source materials while respecting the licensing constraints of real-world music data. 
For human evaluation, participants were informed that their questionnaire responses would be used for research evaluation, and we report only aggregated results without releasing participant-identifying information. 
Since lyric generation may involve unintended imitation of existing copyrighted lyrics, ARIA is intended for research and assisted authoring rather than copying or redistributing existing songs.

\section*{Acknowledgments}

This work was supported by the National Natural Science Foundation of China (No.~62471420)
and the Guangdong Basic and Applied Basic Research Foundation (2025A1515012296).
We also thank the Red Bird MPhil Program at The Hong Kong University of Science and Technology
(Guangzhou) for its funding and resources. This work is partially supported by the Research Travel
Grant of the Base of Red Bird MPhil (RBM) at The Hong Kong University of Science and Technology
(Guangzhou). We would like to thank our RBM Project Supervisor Zheng Luo for the academic support.
\bibliography{custom,4Experiment,appendix_0243_cantonese}

\appendix

\section{Background on Cantonese Tones, the 0243 Method, and Label Derivation}
\label{app:0243}

This appendix gives the background needed to understand why \texttt{0243} is the intermediate representation used in \textsc{ARIA}. Its purpose is to explain why such a collapse is particularly effective for Cantonese lyric writing: unlike English lyric writing, Cantonese must preserve lexical-tone compatibility with melody, and unlike using unreduced citation-tone categories as a control interface, the traditional \texttt{0243} method provides a compact tier sequence that is closer to how Cantonese lyricists reason in practice. We therefore first review basic Cantonese tonal knowledge, the expert origin and practical use of the \texttt{0243} method, and then clarify how our dataset derives final \texttt{0243} labels from aligned lyrics and Jyutping.

\subsection{Cantonese Tone Basics and the Six-Tone Pitch Hierarchy}
\label{app:0243-tones}

Hong Kong Cantonese is commonly analyzed as having six citation tones in open syllables; if checked syllables with stop codas are counted separately, the traditional description is ``nine tones, six pitch categories'' \citep{chao-1947,bauer-benedict-1997}. Tone values are often written in Chao five-level notation, where 1 is the speaker's lowest pitch and 5 the highest \citep{chao-1930}. Exact transcriptions vary, especially for tones 2, 4, and 5, but the relative height relations relevant for lyric writing are stable \citep{khouw-ciocca-2007}. Table~\ref{tab:app-tone-map} summarizes the six Cantonese citation-tone categories, their approximate Chao values, and the corresponding singing tiers and \texttt{0243} labels used in our label derivation.

\begin{table}[t]
\centering
\small
\setlength{\tabcolsep}{0pt}
\renewcommand{\arraystretch}{1.08}
\begin{tabular*}{\columnwidth}{@{\extracolsep{\fill}}ccccc@{}}
\toprule
Tone & Category & Chao & Singing tier & \texttt{0243} \\
\midrule
1 & Yin level      & 55/53 & high     & 3 \\
2 & Yin rising     & 35/25 & high     & 3 \\
3 & Yin departing  & 33    & mid-high & 4 \\
4 & Yang level     & 11/21 & low      & 0 \\
5 & Yang rising    & 13/23 & mid-high & 4 \\
6 & Yang departing & 22    & mid      & 2 \\
\bottomrule
\end{tabular*}
\caption{Mapping Cantonese citation tones to relative singing-pitch tiers and the corresponding \texttt{0243} labels. The exact onset values of tones 2, 4, and 5 vary across descriptions, but the tier relations relevant for \texttt{0243} remain stable \citep{khouw-ciocca-2007,cheng-etal-2025-tonecraft}.}
\label{tab:app-tone-map}
\end{table}

For lyric writing, the key property is therefore not exact spoken contour reproduction, but the robust ordering over tonal height. In this sense, the six-tone system already suggests the four-way tier structure that \texttt{0243} operationalizes.

\subsection{Why the 0243 Method Follows from Cantonese Tone Structure}
\label{app:0243-link}

The \texttt{0243} method does not discard tonal knowledge. It keeps the part that is most useful for melody-constrained lyric writing: relative pitch tier. Cantonese songs preserve tone mainly through ordinal pitch relations among syllables rather than through literal reproduction of spoken F0 trajectories \citep{wong-diehl-2002,schellenberg-gick-2020}. This is why tones with similar relative height in singing can be collapsed without destroying the core tone--melody constraint. Table~\ref{tab:app-strict-relaxed} shows how the stricter nine-tone digit mnemonic collapses into the relaxed \texttt{0243} representation used for lyric writing.

For the six-tone analysis, the relaxed \texttt{0243} mapping can be written as
\begin{equation}
\psi(t) =
\begin{cases}
0, & t = 4,\\
2, & t = 6,\\
4, & t \in \{3,5\},\\
3, & t \in \{1,2\},
\end{cases}
\end{equation}
where \(t\) denotes the Cantonese citation-tone category and \(\psi(t)\) gives its corresponding \texttt{0243} label. If checked tones are retained, tones 7, 8, and 9 collapse together with tones 1, 3, and 6, respectively, because they add coda and duration differences rather than new pitch tiers \citep{chao-1947,bauer-benedict-1997}.

\begin{table}[t]
\centering
\small
\setlength{\tabcolsep}{4pt}
\renewcommand{\arraystretch}{1.08}
\begin{tabularx}{\columnwidth}{@{}c c c >{\centering\arraybackslash}X@{}}
\toprule
Tone & Strict & \texttt{0243} & Rationale \\
\midrule
1 & 3 & 3 & Same high tier as tone 2 \\
2 & 9 & 3 & Same high tier as tone 1 \\
3 & 4 & 4 & Same mid-high tier as tone 5 \\
4 & 0 & 0 & Unique low tier \\
5 & 5 & 4 & Same mid-high tier as tone 3 \\
6 & 2 & 2 & Unique mid tier \\
7 & 1 & 3 & Checked counterpart of tone 1 \\
8 & 8 & 4 & Checked counterpart of tone 3 \\
9 & 6 & 2 & Checked counterpart of tone 6 \\
\bottomrule
\end{tabularx}
\caption{Relation between the stricter nine-tone digit mnemonic and the relaxed \texttt{0243} collapse used in lyric writing. The relaxed form merges categories that occupy the same relative singing-pitch tier \citep{lu-huang-1989-talking,huang-2024-book,huang-2024-intro,cheng-etal-2025-tonecraft}.}
\label{tab:app-strict-relaxed}
\end{table}

This also makes \texttt{0243} musically convenient: it gives lyricists a short discrete sequence that remains grounded in Cantonese tonal phonology. Instead of planning against raw F0 trajectories or full note strings, the lyricist can work with four interpretable tiers: low, mid, mid-high, and high.

\subsection{Expert Origin and Practical Use of the \texttt{0243} Method}
\label{app:0243-expert}

The \texttt{0243} sequence used in \textsc{ARIA} should not be read as a new latent representation introduced by this paper. It is a human expert abstraction from Cantonese lyric-writing practice. Huang describes \texttt{0243} as a ``quasi-scale'' authoring aid: it has scale-like behavior for lyric writing, but is not itself a musical scale \citep{huang-2024-book,huang-2024-intro}. Historically, the method grew out of Cantonese lyric-writing pedagogy in the mid-1980s, was made public in \textit{Talking About Lyric Writing} in 1989, and has since circulated through lyric-writing books, courses, and online tools \citep{lu-huang-1989-talking,huang-2024-intro}. This provenance is important for \textsc{ARIA}: we are not inventing a hidden code and asking users to trust it; we are operationalizing an existing expert interface so that a model can recover it from audio and pass it to a generator.

Public authoring examples show how the method is used. Lyric-writing discussions commonly demonstrate everyday phrases by converting them into short digit strings such as 3322, 3242, or 0230, which trains writers to hear Cantonese words as relative tone-tier patterns \citep{mingpao-2024-0243}. Huang also analyzes real songs by marking a melody, the lyric, and the corresponding \texttt{0243} pattern; one reported production case describes first transcribing a demo into a repeated 202-style digit pattern before writing parallel lyric lines for the song \textit{In Your Eyes} \citep{mingpao-2024-0243}. These examples illustrate why \texttt{0243} is useful as an authoring representation: it is compact, editable, and close to the search process used by human writers.

Academic studies support the same underlying assumption, even when they do not use the \texttt{0243} notation itself. Cantonese tone--tune research tracks concordance between lexical-tone direction and melodic direction in operas and pop songs \citep{li-choi-2016-singing-tones}; large-scale Cantonese pop-song analysis similarly aligns lyrics and melody to quantify tone--melody relations \citep{yang-etal-2021-cantonese-pop}; and computational work on Cantonese lyric generation treats tone--pitch harmony as a central constraint \citep{cheng-etal-2025-tonecraft}. The role of \texttt{0243} in \textsc{ARIA} is therefore both practice-driven and empirically motivated: it is the human-readable tier sequence that connects Cantonese tone knowledge to melody-constrained lexical choice.

\subsection{How We Obtain Character Alignment and \texttt{0243} Labels}
\label{app:0243-manual}

In our data pipeline, the final \texttt{0243} label is not manually inferred from raw audio. Instead, the hard part lies in obtaining reliable character-level alignment from realistic singing audio, after which the final \texttt{0243} sequence is derived deterministically from the lyric-side Jyutping. This choice reflects the practical asymmetry of the problem: once the lyric string is aligned, tone information is already available on the text side, whereas extracting clean timing structure from polyphonic singing remains difficult because accompaniment contaminates the vocal track, sung pitch is unstable under vibrato and glides, and a lyric token does not always map cleanly to a single salient note \citep{wong-etal-2007-lyrics-alignment,fujihara-goto-2012,salamon-etal-2014}.

The present experiments assume that these boundaries are already
available. We did not measure the time required to obtain them or to
prepare alternative symbolic inputs, so we do not claim that boundary
preparation is universally cheaper than constructing a symbolic tone
or melody sequence.

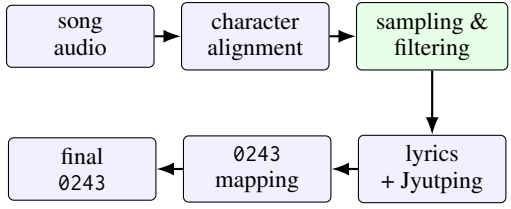
\begin{figure}[t]
\centering
\begin{tikzpicture}[
  >=Latex,
  font=\small,
  box/.style={draw, rounded corners=2pt, align=center, minimum width=1.95cm, minimum height=0.82cm, fill=blue!6},
  human/.style={draw, rounded corners=2pt, align=center, minimum width=2.0cm, minimum height=0.82cm, fill=green!10}
]
\node[box] (audio) {song\\audio};
\node[box, right=0.35cm of audio] (align) {character\\alignment};
\node[human, right=0.35cm of align] (check) {sampling \&\\filtering};
\node[box, below=0.85cm of check] (jyut) {lyrics\\+ Jyutping};
\node[box, left=0.35cm of jyut] (map) {\texttt{0243}\\mapping};
\node[box, left=0.35cm of map] (final) {final\\\texttt{0243}};

\draw[->,thick] (audio) -- (align);
\draw[->,thick] (align) -- (check);
\draw[->,thick] (check) -- (jyut);
\draw[->,thick] (jyut) -- (map);
\draw[->,thick] (map) -- (final);
\end{tikzpicture}
\caption{Practical dataset pipeline used in \textsc{ARIA}. The main quality-control effort is spent on character alignment, while the final \texttt{0243} sequence is derived deterministically from lyric-side Jyutping.}
\label{fig:app-annotation-pipeline}
\end{figure}

\begin{table}[t]
\centering
\small
\setlength{\tabcolsep}{3pt}
\renewcommand{\arraystretch}{1.10}
\begin{tabularx}{\columnwidth}{@{}>{\raggedright\arraybackslash}p{1.65cm}
                                >{\raggedright\arraybackslash}X
                                >{\raggedright\arraybackslash}X@{}}
\toprule
Stage & Main risk & What we do \\
\midrule
Initial alignment
& Pickup notes, melisma, and ornamentation blur where a lyric character should begin and end in the audio.
& We inspect sampled examples and use them to monitor alignment quality instead of assuming the raw alignment output is always reliable. \\

Abnormal-data filtering
& Some examples have durations that are implausibly long, implausibly short, or clearly mismatched with the lyric character count.
& We remove these abnormal items before forming the final dataset so that TRATE is not trained on obvious alignment failures. \\

Jyutping conversion
& The final label sequence depends on having a stable tone category for each aligned lyric character.
& Once the lyrics are aligned, we obtain Cantonese Jyutping on the text side and use it as the basis for deterministic label derivation. \\

\texttt{0243} derivation
& Directly ranking pitch from audio would introduce extra ambiguity and subjective judgment.
& We map Jyutping tone categories to \texttt{0243} deterministically, so no further manual verification of the final \texttt{0243} sequence is required. \\
\bottomrule
\end{tabularx}
\caption{How \textsc{ARIA} obtains alignment-aware \texttt{0243} labels in practice. The pipeline emphasizes alignment quality control and deterministic label derivation rather than manual audio-side ranking of the final labels.}
\label{tab:app-manual}
\end{table}

This design makes the label semantics explicit and reproducible. Figure~\ref{fig:app-annotation-pipeline} summarizes this practical pipeline: the main quality-control effort is placed on character alignment and filtering, while the final \texttt{0243} sequence is deterministically derived from lyric-side Jyutping. Once the character alignment is reliable, the final \texttt{0243} sequence follows from the lyric-side tone information rather than from subjective audio-side judgment. For the learning problem studied in this paper, the main source of label noise therefore comes from alignment failures, which is why sampling and abnormal-data filtering are central parts of dataset construction. Table~\ref{tab:app-manual} further details the main risks at each construction stage and the corresponding quality-control or deterministic derivation step used to obtain alignment-aware \texttt{0243} labels.

\subsection{Dataset Statistics}
\label{app:dataset-stats}

Following the Fei Tsui setting used by ToneCraft \citep{cheng-etal-2025-tonecraft}, we use Cantonese lyrics collected from Fei Tsui as the shared text source for both stages. For Stage~1, these lyrics are paired with matched singing recordings and character-level timestamps, then converted to Jyutping-derived \texttt{0243} labels after alignment filtering. The resulting audio-side resource contains 102,723 aligned lyric--audio line samples from 3,424 recordings. For Stage~2, the Fei Tsui lyric corpus is converted into 171,541 \texttt{0243}-to-lyric pairs. The retrieval resource is built from 0243.hk lexical candidates; enumerating all 2--4 digit \texttt{0243} code patterns and de-duplicating returned strings yields 141,566 unique candidate words. We count 2--4 character candidates because the retrieval module discards single-character entries and uses sliding windows of length 2--4, as described in Appendix~\ref{app:DRA-TCLG-retrieval}.

\begin{figure*}[t]
\centering
\includegraphics[width=\textwidth]{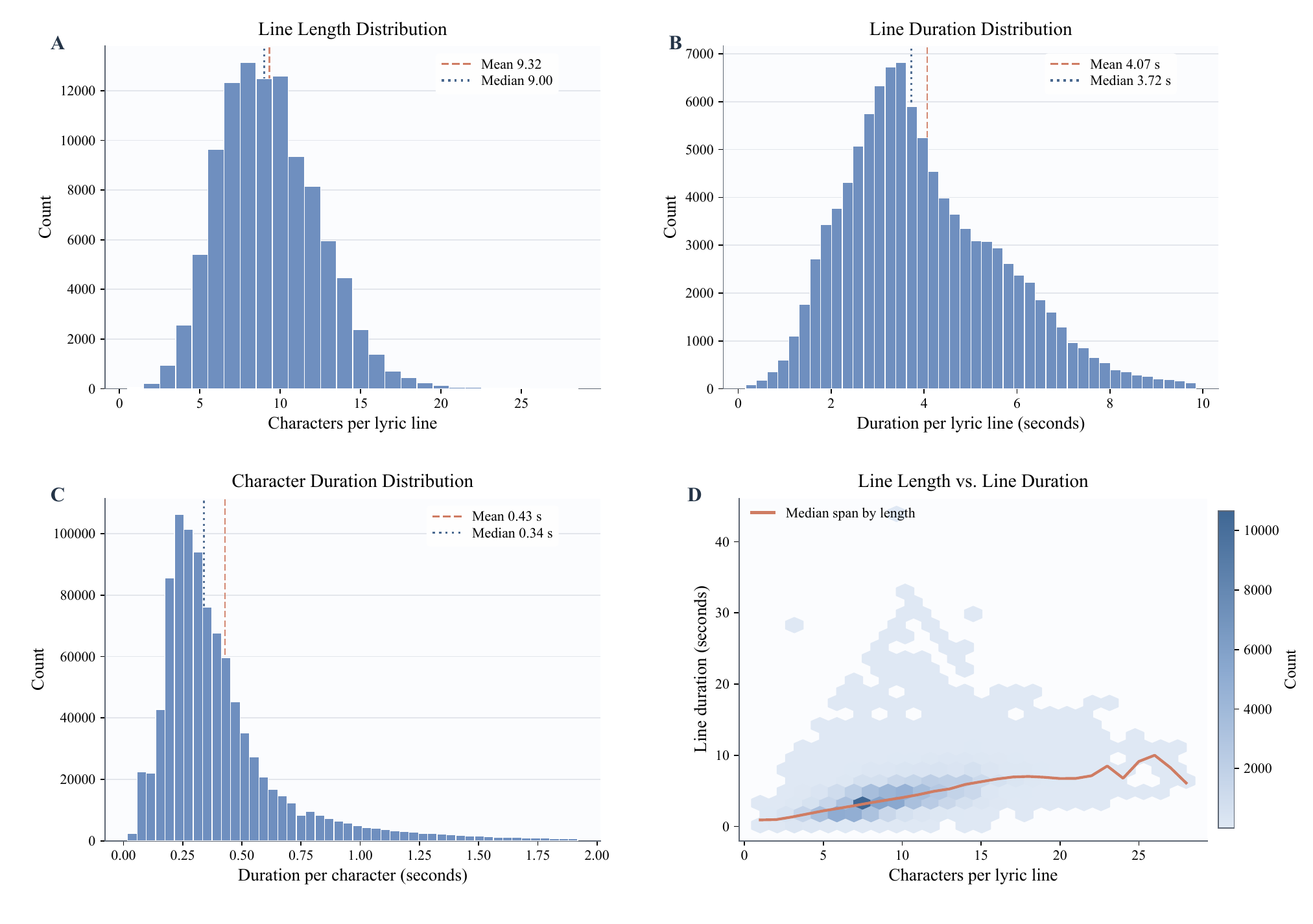}
\caption{Dataset characteristics of \textsc{ARIA}. The figure shows the distribution of lyric-line length, line duration, character duration, and their joint relation. The dataset is dominated by medium-length lines but exhibits a long tail in duration, motivating representations that model both token-local acoustics and sentence-internal relations.}
\label{fig:dataset}
\end{figure*}

Figure~\ref{fig:dataset} provides the detailed distributional view of the \textsc{ARIA} dataset. We place it in the appendix to keep the main experimental section focused on task setup and results, while still documenting the length, duration, and character-timing properties that motivate the TRATE design.

\subsection{Released Annotation Format}
\label{sec:released-annotation-format}
\begin{table}[t]
\centering
\footnotesize
\setlength{\tabcolsep}{0pt}
\renewcommand{\arraystretch}{1.10}
\begin{tabular*}{\columnwidth}{
@{\extracolsep{\fill}}
p{0.32\columnwidth}
p{0.64\columnwidth}
@{}}
\toprule
Field & Description \\
\midrule
\texttt{sample\_id} & Line-level sample identifier \\
\texttt{song\_id} & Source song identifier \\
\texttt{line\_id} & Line index within the song \\
\texttt{text\_length} & Number of lyric characters \\
\texttt{char\_bounds} & Character start time and duration \\
\texttt{label\_0243} & Token-level \texttt{0243} labels \\
\texttt{level4} & Four-level relative height labels \\
\texttt{split} & Train/dev/test split \\
\texttt{title} & Song title, if releasable \\
\texttt{singer} & Singer metadata, if releasable \\
\texttt{jyutping} & Jyutping transcription with tones \\
\texttt{pitches} & Token-level pitch metadata \\
\bottomrule
\end{tabular*}
\caption{Released non-audio annotation format for the Stage 1 audio-to-\texttt{0243} resource. Copyrighted audio files and local experimental paths are not redistributed.}
\label{tab:released-annotation-format}
\end{table}
The released JSON format follows Table~\ref{tab:released-annotation-format}. 
For privacy and portability, the experimental \texttt{audio\_path} field is removed from the public version. The public package instead keeps \texttt{song\_id} and line-level identifiers, which allow researchers with legal access to the source recordings to reconstruct the corresponding audio segments. Fields derived from the lyric text, including Jyutping and \texttt{0243} labels, are released as non-audio annotations for research verification and extension. For compact presentation, \texttt{char\_bounds} denotes the \texttt{character\_boundaries} field in the released JSON files.

\subsection{Why \texttt{0243} Guidance Is Important, Effective, and Efficient}
\label{app:0243-why}

For Cantonese lyric writing, tone is part of lexical meaning rather than an optional prosodic decoration. If melody repeatedly conflicts with the relative tonal tier of the chosen words, the result can hurt both singability and intelligibility \citep{wong-diehl-2002,cheng-etal-2025-tonecraft}. A useful intermediate representation must therefore preserve tone--melody compatibility while remaining compact enough for both human and model use.

\begin{table}[t]
\centering
\small
\setlength{\tabcolsep}{3pt}
\renewcommand{\arraystretch}{1.10}
\begin{tabularx}{\columnwidth}{@{}>{\raggedright\arraybackslash}p{1.55cm}
                                >{\raggedright\arraybackslash}X
                                >{\raggedright\arraybackslash}X@{}}
\toprule
Criterion & Why \texttt{0243} helps & Implication for \textsc{ARIA} \\
\midrule
Important
& It preserves the most rigid Cantonese lyric constraint: compatibility between lexical tone tier and melodic motion.
& The downstream generator does not lose the core ``fit to melody'' condition at the representation boundary. \\

Effective
& It matches long-standing lyric-writing practice and recent harmony-aware Cantonese generation models.
& The control signal is interpretable, linguistically motivated, and aligned with real authoring behavior. \\

Efficient
& Four discrete classes are lower-entropy and more transposition-robust than raw audio, continuous F0, or full note strings.
& Annotation is easier to standardize, search space is smaller, and conditioning an LLM becomes simpler and more stable. \\
\bottomrule
\end{tabularx}
\caption{Why the \texttt{0243} sequence is a strong intermediate control representation for Cantonese lyric authoring.}
\label{tab:app-why0243}
\end{table}

This also explains the design logic of \textsc{ARIA}. A raw audio segment contains much more information than the lyric generator actually needs, while a full symbolic note representation can still be too fine-grained and insufficiently tied to lexical-tone choice. The \texttt{0243} sequence occupies the useful middle layer: compact enough to guide lexical planning, yet rich enough to carry the tone--melody relation that matters for Cantonese songs. Table~\ref{tab:app-why0243} summarizes this motivation from three perspectives: importance for preserving tone--melody compatibility, effectiveness as a practice-aligned authoring representation, and efficiency as a low-entropy control interface. For data-scarce melody-to-lyrics settings, such a low-entropy, human-auditable intermediate representation is also a practical advantage \citep{tian-etal-2023-unsupervised,cheng-etal-2025-tonecraft}.

\subsection{Workflow Alignment of Domain Priors}
\label{app:workflow-priors}

Table~\ref{tab:workflow-priors} summarizes how Cantonese, music, and lyric-writing priors enter the two-stage design. The key point is that the model decomposition mirrors a practical authoring workflow: TRATE performs the listening and tone-tier planning step, while DRA-TCLG performs lexical search and controlled lyric drafting.

\begin{table*}[t]
\centering
\small
\setlength{\tabcolsep}{3pt}
\renewcommand{\arraystretch}{1.10}
\begin{tabularx}{\textwidth}{@{}>{\raggedright\arraybackslash}p{2.45cm}
                                >{\raggedright\arraybackslash}X
                                >{\raggedright\arraybackslash}X
                                >{\raggedright\arraybackslash}X@{}}
\toprule
Real-world lyric-writing step
& Prior knowledge used by human lyricists
& Corresponding \textsc{ARIA} design
& Effect on the model \\
\midrule
Listen to a demo and locate singable slots
& Musical timing, phrase boundaries, and character-level rhythm.
& TRATE takes singing audio with character boundaries $B=\{(s_i,e_i)\}_{i=1}^{N}$.
& The model predicts one controllable label per lyric position rather than a frame-level F0 trace. \\

Estimate tone-compatible melody tiers
& Cantonese tones are organized by relative pitch height and contour, not only by absolute frequency.
& TRATE uses absolute, contour, and line-internal relative feature streams.
& Acoustic, melodic, and sentence-level evidence are separated before fusion. \\

Judge neighboring tone--melody fit
& Lyricists compare adjacent syllables and avoid implausible large jumps when possible.
& TRATE uses relation-aware attention and pairwise ordering heads to model short-range melodic relations.
& The target is treated as relational, matching the ordinal nature of \texttt{0243}. \\

Use \texttt{0243} as a draftable authoring plan
& The expert \texttt{0243} method collapses Cantonese tones into four practical singing tiers.
& TRATE predicts level-4 labels and maps them deterministically to \texttt{0243}; structured heads regularize register and contour factors.
& Cantonese tonal knowledge enters supervision without giving the model lyric text at inference time. \\

Search and revise lexical candidates
& Lyricists consult tone-compatible dictionaries and allow near alternatives for content.
& DRA-TCLG conditions on $P$, retrieves from a \texttt{0243}--lyrics dictionary, and refines candidates with an LLM.
& The generator receives an explicit, human-readable lexical-planning signal instead of an opaque audio embedding. \\
\bottomrule
\end{tabularx}
\caption{How Cantonese, music, and lyric-writing priors are integrated into \textsc{ARIA}. TRATE corresponds to listening and tone-tier planning; DRA-TCLG corresponds to lexical search and controlled lyric drafting.}
\label{tab:workflow-priors}
\end{table*}

\subsection{Why Audio-to-\texttt{0243} Mapping Is a Distinct Task}
\label{app:audio-to-0243}

TRATE's task is not simply melody extraction, lyrics alignment, or melody-to-lyrics generation under a new name. Melody extraction estimates frame-level or note-level pitch from audio; lyrics alignment locates text in time; symbolic melody-to-lyrics systems consume a score-like melody or a note sequence as the generation condition \citep{sheng-etal-2021-songmass,tian-etal-2023-unsupervised,miyano-saito-2021-vocal-melody}. Audio-to-\texttt{0243} mapping instead asks for a token-level authoring sequence: given realistic singing audio and character boundaries, recover the four-tier pattern that a Cantonese lyricist would use to guide lexical choice. This target is discrete, relative, and line-dependent, so it cannot be obtained by directly thresholding absolute F0.

This task is necessary because many real authoring settings begin with a demo, recording, or hummed melody rather than a clean symbolic score. A lyricist can listen to such audio and infer an approximate \texttt{0243} plan before choosing words; a model-based system needs the same front end if it is to support audio-driven lyric authoring. Prior Cantonese studies have shown that extracting and aligning singing voices from polyphonic tracks is feasible for analysis, but still challenging because accompaniment, ASR errors, and character-timing ambiguity affect the recovered melody--lyric relation \citep{wong-etal-2007-lyrics-alignment,yang-etal-2021-cantonese-pop,salamon-etal-2014}. TRATE turns this difficult audio-side evidence into the exact form needed by DRA-TCLG: a compact, editable, and dictionary-searchable control sequence.

The utility of the task is also broader than the current DRA-TCLG pipeline. A reliable audio-to-\texttt{0243} mapper can support automatic preparation of lyric-writing prompts from recordings, semi-automatic checking of whether candidate Cantonese lyrics fit a melody, retrieval of tone-compatible lexical candidates, educational tools for Cantonese songwriting, and corpus analysis of tone--melody practice. To our knowledge, prior work has not provided a dedicated benchmark for recovering \texttt{0243}-style lyric-authoring guidance directly from realistic Cantonese singing audio. We therefore frame TRATE as a new and necessary bridge between MIR-style audio analysis and controllable Cantonese lyric generation.

\subsection{Why Within \texorpdfstring{$\pm1$}{+-1} Is a Music-Aware Metric}
\label{app:withinpmone}

Standard music-information-retrieval evaluation rarely treats exact pitch agreement as the only meaningful notion of correctness. In predominant-melody and pitch-tracking tasks, nearby estimates are commonly treated as acceptable under tolerance-aware metrics, and contour- or chroma-oriented measures are reported alongside exact scores because small local deviations may preserve the musically salient identity of the melodic event \citep{salamon-etal-2014,raffel-etal-2014-mireval}. In practice, common melody-evaluation protocols already score frame-level correctness with pitch tolerances rather than exact-frequency identity, which makes our Within $\pm1$ metric a natural analogue after collapsing the target into four ordinal singing tiers \citep{salamon-etal-2014,raffel-etal-2014-mireval}. More generally, melody-perception research shows that contour and relative motion remain robust cues for melody recognition even when exact pitch intervals are not preserved perfectly \citep{dowling-1978}.

This logic fits our setting particularly well. The target space in TRATE is already a coarse four-way ordinal collapse rather than a dense absolute-pitch scale, and work on Cantonese singing argues that lexical tone is preserved mainly through relative pitch relations rather than literal reproduction of spoken F0 contours \citep{wong-diehl-2002,schellenberg-gick-2020}. Therefore, a one-tier error is a musically plausible near miss: it still lands in an adjacent singing tier, preserves local neighborhood structure, and usually keeps the prediction close to the control granularity that DRA-TCLG actually consumes, even though it fails exact classification. This last point about DRA-TCLG is an inference from our task design, while the pitch-proximity argument follows from the cited tone-singing and MIR literature. By contrast, larger errors jump across nonadjacent tiers and are more likely to alter the intended melody--tone fit. For this reason, Within $\pm1$ complements exact token accuracy and Macro-F1 rather than replacing them.

\section{Additional Experimental Setup and Result Analysis}
\label{app:exp-details}

\subsection{Experimental Setup Details}
\label{app:exp-setup-details}

TRATE takes singing audio and character timestamps as input and predicts token-level \texttt{0243} labels. We report four complementary metrics. \textbf{TokAcc} measures exact token-level recovery; \textbf{Macro-F1} emphasizes balanced performance across labels; \textbf{Within $\pm1$} measures whether the predicted underlying four-way height is at most one level away from the reference; and \textbf{NED} denotes normalized edit distance over the predicted \texttt{0243} sequence. Together, these metrics cover exact correctness, label balance, tolerance-aware musical proximity, and sequence usability for downstream control.

The audio-to-\texttt{0243} comparison includes two parameter-free quartile heuristics and two learned Qwen3.5 baselines~\citep{qwen3.5}. 
For the quartile heuristics, we compute one scalar pitch value for each character in a lyric line, calculate the line-level 25th, 50th, and 75th percentile thresholds, and assign characters to four relative height bins from low to high. 
The resulting height labels are then mapped to \texttt{0243} using $g(1)=0$, $g(2)=2$, $g(3)=4$, and $g(4)=3$. 
Rule RQ applies this procedure to \texttt{zscore\_median}, the character median log-F0 normalized by the line mean and standard deviation, while Rule AQ applies it to \texttt{mean\_log\_f0}, the absolute mean log-F0 inside the character window. 
These rules test whether simple line-relative pitch ranking is sufficient for recovering the \texttt{0243} plan.

For the learned baselines, we fine-tune Qwen3.5 with LoRA as a supervised sequence predictor. 
Each instance serializes the audio-side feature sequence into a compact text prompt, and the target output is the corresponding token-level \texttt{0243} sequence. 
Qwen Rel. receives only relative features, including line-normalized height, percent rank, neighboring pitch intervals, local peak/valley flags, position, and relative duration. 
Qwen 3-Br. receives the full three-stream representation used by TRATE, including absolute statistics, resampled contour features, and line-internal relative features. 
Both baselines use the same train/dev/test split as TRATE, supervise only the \texttt{0243} labels, and decode deterministically at inference time without access to lyric text, Jyutping strings, or symbolic pitch labels.

DRA-TCLG uses Qwen3.5-4B~\citep{qwen3.5} with LoRA
adapters~\citep{hu2022lora} on the attention projection layers.
The backbone is frozen. We use rank $r=16$, scaling factor
$\alpha=32$, dropout 0.05, bf16 precision, three epochs for base SFT
at learning rate $1\times10^{-5}$, and two epochs for
retrieval-guided adaptation at $5\times10^{-6}$.

Our ToneCraft-Qwen2-7B baseline is an in-domain reproduction of
ToneCraft~\citep{cheng-etal-2025-tonecraft}. We retain the
Qwen2-7B backbone and the tone-aware, character-level generation
setting of the original method, while training and evaluating it on
our Fei Tsui \texttt{0243}-to-lyric pairs so that it faces the same
data distribution as DRA-TCLG. Retaining the original backbone avoids
altering the baseline architecture solely to match the backbone used
by our model.

We also follow ToneCraft's comparison protocol when adapting SmBART and SongNet. SmBART is based on scansion-based lyric generation \citep{chen-teufel-2024-scansion}; following ToneCraft, we adapt its three-region contour control to the four-region Cantonese tone setting by merging high and mid-high regions where the baseline requires a three-way representation. SongNet \citep{li-etal-2020-rigid} is included as a rigid-format lyric-generation baseline; following ToneCraft, it is trained with fixed title/format controls and the same tonal-region information used for the SmBART-style comparison. These baselines test whether existing contour-control or rigid-format generators can use the \texttt{0243} condition as effectively as a generator trained directly around this interface.

Following ToneCraft \citep{cheng-etal-2025-tonecraft}, we evaluate generated lyrics from the perspectives of lexical diversity and \texttt{0243}--lyric alignment. Lexical similarity is measured by encoding generated lyrics using \texttt{text2vec-base-chinese} \citep{xu-2022-text2vec} and computing pairwise cosine similarities, from which average similarity (\textbf{Avg Sim}) and minimum similarity (\textbf{Min Sim}) are reported; lower values indicate higher diversity. We further report \textbf{MaD1/MaD2} and \textbf{MiD1/MiD2}, computed from unigram and bigram entropy statistics. For alignment, \textbf{Consistency} measures rank correlation between lyric tone contours and melody pitch contours, while \textbf{Harmony} evaluates compatibility under Cantonese tone--pitch correspondence constraints.

\subsection{Main Result Interpretation}
\label{app:main-result-details}

The audio-to-\texttt{0243} results show a clear hierarchy. The two rule systems reach only about 0.37 TokAcc and 0.36 Macro-F1, indicating that coarse quartile heuristics capture only weak regularities. Switching to Qwen3.5 yields a substantial gain, and conditioning on all three feature groups is clearly better than using relative features alone. However, the best learned baseline still falls short of TRATE. The gain is not limited to exact token prediction: TRATE also achieves the best Within $\pm1$ score and the lowest NED. Because DRA-TCLG consumes the whole predicted \texttt{0243} sequence rather than isolated labels, the NED reduction directly improves the downstream control interface.

The \texttt{0243}-to-lyric results suggest that the main challenge is not generating diverse Chinese lyrics, but constraining lexical choice under a tone--melody plan. 
Models such as SongNet and SmBART are expressive enough to produce varied surface forms, yet their diversity is only weakly tied to the \texttt{0243} condition. 
As a result, diversity-oriented metrics can improve even when the generated words violate the intended Cantonese tonal tiers. 
ToneCraft-Qwen2 shows that explicit tone-aware modeling is a strong inductive bias, but its remaining gap to DRA-TCLG indicates that tone awareness alone does not fully solve controllable lexical selection. 
DRA-TCLG benefits from two additional constraints: the generator is trained with the same \texttt{0243} sequence used at test time, and retrieval exposes candidate words that already satisfy local tone patterns. 
Because these candidates are provided as soft lexical priors rather than hard constraints, the model can still choose fluent alternatives when retrieved words are semantically inappropriate. 
This explains the observed balance between high Harmony/Consistency and competitive diversity scores.

\subsection{Representative Error-Propagation Cases}
\label{app:error-cases}

The Jyutping transcriptions below were generated using the
project environment's \texttt{ToJyutping} 3.2.0 converter.

\paragraph{Single adjacent-tier substitution.}
The reference lyric is
\zh{仍然是愛你，一生都只會這樣說}.
Its Jyutping transcription is
\jyut{jing4 jin4 si6 oi3 nei5, jat1 sang1 dou1 zi2 wui5 ze2 joeng6 syut3},
and its English gloss is
``I still love you; throughout my life, I would only say it this
way.'' The lyric-derived reference plan is
\texttt{0024433334424}.

TRATE predicts \texttt{0224433334424}, and DRA-TCLG generates
\zh{才學會愛你的方式因你太倔強}.
Its Jyutping transcription is
\jyut{coi4 hok6 wui2 oi3 nei5 dik1 fong1 sik1 jan1 nei5 taai3 gwat6 koeng5},
and its English gloss is
``Only then did I learn how to love you, because you are too
stubborn.'' The extracted output plan is
\texttt{0224433334424}, yielding final-vs-reference
Harmony/Consistency of 0.9514/0.9909. The adjacent-tier substitution
is therefore not corrected in code space, but its downstream effect
remains limited.

\paragraph{Eight-position error run.}
The reference lyric is
\zh{誰人在舞，誰人在唱}.
Its Jyutping transcription is
\jyut{seoi4 jan4 zoi6 mou5, seoi4 jan4 zoi6 coeng3},
and its English gloss is
``Who is dancing, and who is singing?'' The lyric-derived reference
plan is \texttt{00240024}.

TRATE instead predicts \texttt{44334433}, and DRA-TCLG generates
\zh{愛過總因太怕分開}.
Its Jyutping transcription is
\jyut{oi3 gwo3 zung2 jan1 taai3 paa3 fan1 hoi1},
and its English gloss is
``We loved, yet were always too afraid to part.'' The extracted
output plan is exactly \texttt{44334433}. Its
final-vs-predicted-plan Harmony/Consistency/Match scores are
1.0000/1.0000/1.0000, whereas its final-vs-reference scores fall to
0.1935/0.9428/0.0000. This case shows that Stage~2 introduces no
additional plan mismatch, but it cannot recover from a severely
incorrect Stage~1 plan.

\paragraph{Category-level error profile.}
Exact accuracies for labels 0/2/4/3 are 0.8147/0.6923/0.7744/0.8634,
respectively, making label~2 the most difficult class. Label~0 has the highest
non-adjacent-error rate (0.0998). These findings identify confusions involving the
difficult label~2, non-adjacent substitutions from label~0, and consecutive error
runs as the main remaining Stage~1 failure modes. They motivate ordinal- and
class-aware losses, sequence-level penalties for error runs, and confidence-based
local re-prediction as future directions rather than measured improvements.

\subsection{Ablation Details}
\label{app:ablation-details}

The TRATE ablation clarifies the role of each component. Moving from a rel-only Transformer to the three-branch representation yields a large TokAcc gain and NED reduction, showing that absolute and contour cues add information beyond line-internal rank alone. Adding structured relation-aware modeling further improves both exact and sequence-level metrics, supporting the claim that the target is inherently relational rather than purely local. Token-wise gated fusion provides the final improvement, indicating that the most reliable evidence source varies across tokens.

The DRA-TCLG ablation shows that retrieval must be introduced carefully. Compared with the Base model, directly applying retrieval augmentation in a one-stage manner sharply degrades Harmony and Consistency, indicating that noisy candidate lists interfere with stable tone-to-lyric alignment learning. The diversity columns explain the trade-off: One-Stage RAG increases several diversity scores, but it does so while breaking alignment, which is unacceptable for Cantonese lyric writing. Two-Stage RAG combines the advantages of both settings. The base model first learns how sequence length and code pattern constrain a lyric line, and the second phase adds lexical evidence after this mapping is stable. Retrieval is therefore treated as optional authoring context rather than a hard decoding constraint.

\subsection{Boundary Perturbation Robustness}
\label{app:boundary-robustness}

Since TRATE assumes character-level boundaries, we evaluate whether
its audio-side features are overly sensitive to boundary errors. This
experiment is designed as a within-checkpoint sensitivity analysis,
rather than another estimate over training randomness. We fix the best
EMA checkpoint from training seed 3407 throughout the experiment.

For each perturbation magnitude
$r \in \{25,50,100\}$ ms, we independently sample start and end
offsets uniformly within $\pm r$ ms, recompute all audio-side
features, and run deterministic inference. Each nonzero perturbation
condition reports the mean over three perturbation seeds. The clean condition uses zero offsets and is evaluated once through
the same feature-recomputation pipeline.\footnote{The recomputed clean
prediction differs from the original cached seed-3407 evaluation at
only one of 94,008 tokens because the acoustic features are rebuilt
through the robustness pipeline.} Accordingly, its value is not the
three-training-seed average reported in Table~\ref{tab:main-results}.

Table~\ref{tab:boundary-robustness} shows that performance remains
stable under $\pm25$ ms perturbation. At $\pm50$ ms, Acc and F1
decrease by 0.0068 and 0.0071, respectively, while Within $\pm1$
changes by only 0.0003 and NED increases by 0.0053. Under
$\pm100$ ms perturbation, the degradation is clearer: Acc and F1
decrease by 0.0281 and 0.0302, respectively, while Within $\pm1$
remains relatively high at 0.8773. These results indicate that
accurate character boundaries are beneficial, but the fixed TRATE
model remains moderately robust to realistic boundary offsets, with
many residual errors remaining in musically adjacent tiers.

\begin{table}[t]
\centering
\small
\setlength{\tabcolsep}{0pt}
\renewcommand{\arraystretch}{1.08}
\begin{tabular*}{\columnwidth}{@{\extracolsep{\fill}}lcccc@{}}
\toprule
Noise & Acc $\uparrow$ & F1 $\uparrow$ &
$\pm1$ $\uparrow$ & NED $\downarrow$ \\
\midrule
Clean
& \textbf{0.6459}
& \textbf{0.6173}
& \underline{0.8840}
& \textbf{0.3462} \\
$\pm25$ ms
& \underline{0.6447}
& \underline{0.6167}
& \textbf{0.8853}
& \underline{0.3465} \\
$\pm50$ ms
& 0.6391
& 0.6102
& 0.8837
& 0.3515 \\
$\pm100$ ms
& 0.6178
& 0.5871
& 0.8773
& 0.3704 \\
\bottomrule
\end{tabular*}
\caption{Boundary-perturbation robustness of the fixed seed-3407
EMA checkpoint. Clean is evaluated once using the same
feature-recomputation pipeline as the perturbed conditions; each
nonzero noise condition reports the mean over three perturbation
seeds.}
\label{tab:boundary-robustness}
\end{table}

\section{Feature Preprocessing Details}
\label{app:feature-details}

All baselines and TRATE use the same audio-side preprocessing backbone. We first extract frame-level F0, aggregate it within each character boundary, and then construct three complementary feature streams. The absolute stream summarizes token-local acoustic and timing statistics before line-level normalization (Table~\ref{tab:absolute-features}); the contour stream preserves the resampled within-character F0 trajectory and its pointwise reliability indicators (Table~\ref{tab:contour-features}); and the relative stream encodes line-internal pitch height, ordinal position, neighborhood relations, and relative timing from token-level median log-F0 and duration (Table~\ref{tab:relative-features}). These three tables list the branch inputs used in the reported experiments, with cache-only aliases and backward-compatible duplicate names omitted.

\begin{table*}[t]
\centering
\small
\setlength{\tabcolsep}{4pt}
\renewcommand{\arraystretch}{1.08}
\begin{tabularx}{\textwidth}{@{}L{3.4cm}YY@{}}
\toprule
Feature & Definition & Role \\
\midrule
\texttt{mean\_log\_f0} & Mean log-F0 over frames assigned to the character. & Direct token-window pitch height before comparison with other tokens. \\
\texttt{median\_log\_f0} & Median log-F0 over the character window. & Robust absolute height summary for the token itself. \\
\texttt{std\_log\_f0} & Standard deviation of log-F0 in the character window. & Local acoustic dispersion, not normalized by line context. \\
\texttt{min\_log\_f0} & Minimum log-F0 in the character window. & Lower bound of the token's observed pitch region. \\
\texttt{max\_log\_f0} & Maximum log-F0 in the character window. & Upper bound of the token's observed pitch region. \\
\texttt{q25\_log\_f0} & 25th percentile of character-window log-F0. & Absolute lower-quartile pitch statistic within the token. \\
\texttt{q75\_log\_f0} & 75th percentile of character-window log-F0. & Absolute upper-quartile pitch statistic within the token. \\
\texttt{start\_end\_delta} & Last minus first log-F0 value inside the character window. & Direct within-token movement before line-level relational normalization. \\
\texttt{voiced\_ratio} & Fraction of frames in the character window marked voiced. & Token-local F0 reliability and voicing evidence. \\
\texttt{duration} & Character duration in seconds. & Raw timing evidence for the character slot. \\
\texttt{linear\_slope} & Linear-fit slope of log-F0 over the character window. & Direct local trend in the token's pitch trace. \\
\texttt{quadratic\_coef} & Quadratic coefficient fitted to the token log-F0 trace. & Direct local curvature before any line-wise comparison. \\
\texttt{pitch\_range} & \texttt{max\_log\_f0} minus \texttt{min\_log\_f0}. & Token-internal pitch span from absolute acoustic values. \\
\texttt{voiced\_frame\_count} & Number of voiced frames inside the character window. & Token-local quantity indicating available native F0 evidence. \\
\texttt{valid\_f0\_mask} & Indicator that the character window has at least one native voiced frame. & Token-local validity flag for the absolute F0 summary. \\
\texttt{filled\_f0\_mask} & Indicator that missing F0 had to be filled or interpolated. & Token-local missingness flag tied to the absolute F0 summary. \\
\bottomrule
\end{tabularx}
\caption{Absolute acoustic features. These features summarize token-local F0 and timing evidence before line-level normalization or cross-token comparison.}
\label{tab:absolute-features}
\end{table*}

\begin{table*}[t]
\centering
\small
\setlength{\tabcolsep}{4pt}
\renewcommand{\arraystretch}{1.08}
\begin{tabularx}{\textwidth}{@{}L{3.8cm}YY@{}}
\toprule
Feature & Definition & Role \\
\midrule
\texttt{contour\_log\_f0}[0:15] & Sixteen evenly resampled log-F0 points across the character window. & Preserves pointwise within-character shape rather than collapsing the token into scalar statistics. \\
\texttt{contour\_voiced\_mask}[0:15] & Sixteen pointwise voiced/unvoiced indicators aligned to the resampled contour. & Attaches reliability evidence to each contour point. \\
\texttt{contour\_filled\_mask}[0:15] & Sixteen pointwise indicators showing where contour values were imputed. & Marks missing or filled trajectory regions in the contour sequence. \\
\texttt{token\_valid\_voiced} & Indicator that the character window contains native voiced frames before filling. & Provides a summary reliability flag for the extracted contour trajectory. \\
\bottomrule
\end{tabularx}
\caption{Contour features. These features encode the within-character F0 trajectory and its pointwise reliability.}
\label{tab:contour-features}
\end{table*}

\begin{table*}[t]
\centering
\small
\setlength{\tabcolsep}{4pt}
\renewcommand{\arraystretch}{1.08}
\begin{tabularx}{\textwidth}{@{}L{3.5cm}YY@{}}
\toprule
Feature & Definition & Role \\
\midrule
\texttt{robust\_z\_median} & Character median log-F0 minus line median, divided by the line interquartile range. & Defines height relative to the current lyric line. \\
\texttt{zscore\_median} & Character median log-F0 minus line mean, divided by line standard deviation. & Provides line-normalized height; this is the scalar used by Rule RQ. \\
\texttt{percent\_rank\_median} & Stable percent rank of character median log-F0 among tokens in the line. & Captures ordinal line-internal height independent of absolute register. \\
\texttt{prev\_interval} & Current median log-F0 minus previous-token median log-F0. & Encodes adjacent-token pitch relation to the left context. \\
\texttt{next\_interval} & Next-token median log-F0 minus current median log-F0. & Encodes adjacent-token pitch relation to the right context. \\
\texttt{prev\_higher\_flag} & Indicator that the current token is higher than the previous token. & Represents directional relation between neighboring tokens. \\
\texttt{next\_higher\_flag} & Indicator that the next token is higher than the current token. & Represents directional relation between neighboring tokens. \\
\texttt{local\_peak\_flag} & Indicator that the current token is a local peak against both neighbors. & Marks a local contour role based on neighboring-token comparison. \\
\texttt{local\_valley\_flag} & Indicator that the current token is a local valley against both neighbors. & Marks a local contour role based on neighboring-token comparison. \\
\texttt{absolute\_position} & Zero-based token index inside the lyric line. & Encodes the token's line-internal sequence coordinate. \\
\texttt{normalized\_position}
& Zero-based token index divided by \(N-1\) for \(N>1\), and 0 for
\(N=1\).
& Provides length-invariant position within the current line. \\
\texttt{final\_lengthening\_flag} & Indicator that the final token is long relative to the line median duration. & Combines line position with duration relative to the same line. \\
\texttt{relative\_duration} & Character duration divided by the line median character duration. & Expresses duration against the current line's timing scale. \\
\bottomrule
\end{tabularx}
\caption{Relative features. These features normalize pitch, duration, and position within each lyric line and encode local cross-token relations.}
\label{tab:relative-features}
\end{table*}

\section{Mandarin-to-Cantonese Transfer Stress Test}
\label{app:crosslingual-stress}

Although TRATE is trained and evaluated primarily in the in-domain Cantonese setting, we additionally test cross-lingual transfer from Mandarin source clips to Cantonese \texttt{0243} targets. We use paired songs for which a Mandarin version precedes a corresponding Cantonese version. Each model receives the Mandarin-version singing audio together with character-level timestamps, and its output is evaluated against the \texttt{0243} sequence derived from the Cantonese lyrics. This setting is substantially harder than the main benchmark because the source and target lyric lengths may differ, token correspondence is imperfect, and adapted lyrics may redistribute syllables across similar melodic phrases. We therefore view this experiment as a generalization analysis and stress test rather than as the primary supervised evaluation.

\begin{table}[t]
\centering
\small
\setlength{\tabcolsep}{0pt}
\renewcommand{\arraystretch}{1.08}
\begin{tabular*}{\columnwidth}{@{\extracolsep{\fill}}lcccc@{}}
\toprule
Model & Acc $\uparrow$ & F1 $\uparrow$ & $\pm1$ $\uparrow$ & NED $\downarrow$ \\
\midrule
Rule RQ    & 0.399 & 0.436 & 0.593 & 0.625 \\
Rule AQ    & 0.394 & 0.430 & 0.608 & 0.626 \\
Qwen Rel.  & \textbf{0.460} & \textbf{0.489} & 0.671 & \textbf{0.544} \\
Qwen 3-Br. & 0.394 & 0.409 & 0.658 & 0.614 \\
TRATE     & \underline{0.454} & \underline{0.471} & \textbf{0.677} & \underline{0.550} \\
\bottomrule
\end{tabular*}
\caption{Mandarin-to-Cantonese transfer stress test. Models take Mandarin-version audio with character-level timestamps as input, while references are \texttt{0243} sequences derived from the corresponding Cantonese lyrics. Acc, F1, and $\pm1$ denote TokAcc, Macro-F1, and Within $\pm1$. Higher is better except for NED.}
\label{tab:crosslingual-stress}
\end{table}

Table~\ref{tab:crosslingual-stress} shows that exact cross-version prediction is difficult for all systems. Qwen Rel. achieves the best Acc, F1, and NED, indicating that a large-model baseline remains strong when sequence correspondence is noisy. TRATE nevertheless remains competitive: its NED is close to Qwen Rel. (0.550 vs. 0.544), it improves over Qwen 3-Br. on every metric, and it obtains the best Within $\pm1$ score. Because Within $\pm1$ measures adjacent-tier musical proximity, this result suggests that TRATE preserves the ordinal melody--tone relation even when exact Mandarin-to-Cantonese token correspondence is unstable. We also observe fewer length-mismatch failures for TRATE outputs in this setting, which supports its role as a robust control-sequence producer under adaptation stress.

\section{Additional TRATE Model and Experimental Details}
\label{app:stage1-extra}

\subsection{Structured Auxiliary Heads and Training Objective}
\label{app:stage1-details}

On top of the contextualized representation $z_i$, TRATE uses one main prediction head and several structured auxiliary heads.

\paragraph{Height head.}
The main head predicts the four-way relative pitch distribution:
\[
p_i^{h} = \mathrm{softmax}(W_h z_i + b_h).
\]

\paragraph{Contour-family and register heads.}
We additionally predict a contour-family distribution $p_i^{c}$ and a register distribution $p_i^{r}$. Rather than treating the traditional six-tone system as a flat label space, we factor part of the tonal supervision through these two components, form a structured intermediate distribution $p_i^{t}=\mathrm{reshape}(p_i^{r}\otimes p_i^{c})$, and map it to a derived height distribution:
\[
\tilde{p}_i^{h}=M_{t\rightarrow h}^{\top}p_i^{t}.
\]

\paragraph{Pairwise relation head.}
We also predict short-range pairwise ordering relations over token pairs
\[
\mathcal{P}=\{(i,j)\mid j-i\in\{1,2\}\}.
\]
For each pair in $\mathcal{P}$, the model predicts whether the underlying level-4 target of token $i$ is lower than, equal to, or higher than that of token $j$.

The overall objective combines the main height loss, auxiliary classification losses, and two regularizers:
\begin{equation}
\begin{aligned}
\mathcal{L}
= {}&
\lambda_h \mathcal{L}_{\mathrm{height}}
+\lambda_c \mathcal{L}_{\mathrm{contour}}
+\lambda_r \mathcal{L}_{\mathrm{register}} \\
&+\lambda_t \mathcal{L}_{\mathrm{tone-6}}
+\lambda_p \mathcal{L}_{\mathrm{pair}}
+\lambda_{\mathrm{hc}} \mathcal{L}_{\mathrm{height\text{-}cons}} \\
&+\lambda_{\mathrm{tc}} \mathcal{L}_{\mathrm{trans\text{-}cons}},
\end{aligned}
\end{equation}
where \(\lambda_h,\lambda_c,\lambda_r,\lambda_t,\lambda_p,\lambda_{hc}\), and \(\lambda_{tc}\) are non-negative loss weights. 
\(\mathcal{L}_{\mathrm{height}}\) is the label-smoothed cross-entropy loss for the main level-4 height prediction; 
\(\mathcal{L}_{\mathrm{contour}}\) and \(\mathcal{L}_{\mathrm{register}}\) supervise the contour-family and register heads; 
\(\mathcal{L}_{\mathrm{tone\text{-}6}}\) is computed from the structured tone distribution \(p_i^t\), rather than from an independent flat tone-6 head; 
\(\mathcal{L}_{\mathrm{pair}}\) supervises pairwise ordering relations; 
\(\mathcal{L}_{\mathrm{height\text{-}cons}}\) regularizes agreement between \(p_i^h\) and \(\tilde{p}_i^h\); 
and \(\mathcal{L}_{\mathrm{trans\text{-}cons}}\) enforces consistency under pitch-shift augmentation.
The main loss is cross-entropy with label smoothing \citep{szegedy2016rethinking}. The first regularizer enforces agreement between the direct height head and the height distribution derived from the structured tonal pathway, while the second enforces transposition consistency under pitch-shift augmentation. Together, these terms improve both token-level prediction and line-level sequence stability.

\section{DRA-TCLG Input Formats and Retrieval Details}
\label{app:DRA-TCLG-retrieval}

This section describes the input formats and retrieval rules used by DRA-TCLG.

\subsection{Phase 1: Base Tone-to-Lyric Input}

In the first DRA-TCLG phase, the model is trained without retrieval augmentation. Each sample consists of a tone sequence $x$ and a target lyric $y$. The input is formatted as:

\begin{quote}
\textbf{Instruction:} \textit{Generate Cantonese lyrics from the following 0243 sequence. Output only one line of lyrics without explanation.} \\
\textbf{0243 sequence:} \{P\}
\end{quote}

No external context is provided. The model learns a direct mapping from $x$ to $y$ using only the tone-conditioned generation signal.

\subsection{Phase 2: Retrieval-Augmented Input}

In the second DRA-TCLG phase, structured retrieval context $r$ is introduced to provide lexical priors for generation. The input format is:

\begin{quote}
\textbf{Instruction:} \textit{Generate Cantonese lyrics from the following 0243 tone sequence. Output only one line of lyrics without explanation.} \\
\textbf{Retrieved lexical candidates:} 
\textit{Please prioritize the following 0243 candidate words when generating lyrics. Each line follows the format "0243 code: candidate words". Try to use candidate words that naturally match both the current code pattern and the semantic context. If a candidate does not fit fluently, it may be ignored.}

\begin{quote}
\small
\ttfamily
0243\_code : candidate\_words \\
44 : candidate\_1, candidate\_2, candidate\_3, candidate\_4 \\
42 : candidate\_1, candidate\_2, candidate\_3, candidate\_4 \\
\ldots
\end{quote}

\textbf{0243 sequence:} \{P\}
\end{quote}

Here, $r$ denotes retrieval results grouped by \texttt{0243} code to preserve local alignment between tone segments and candidate words. The final input is formulated as:
\begin{equation}
p_{\text{rag}} = [\text{instruction}; r; P].
\end{equation}

The retrieval context is used during both training and inference. The loss is computed only over target lyric tokens, while the instruction, tone sequence, and retrieval context are fully masked.
\subsection{Retrieval-Augmented Candidate Construction}

\paragraph{Frequency-aware \texttt{0243} dictionary.}

The retrieval module is built on 0243.hk lexical resources. To improve retrieval relevance, we compute the occurrence frequency of each candidate word from the Fei Tsui training lyrics and construct a frequency-aware dictionary:
\begin{equation}
D = \{c \rightarrow (w_1, w_2, \dots)\},
\end{equation}
where candidate words under each \texttt{0243} code are sorted according to their frequency in real lyrics.

Compared with the original dictionary, the proposed version preserves the same mapping structure while prioritizing high-frequency lyrical expressions. This strategy improves the quality of lexical priors and reduces the probability of retrieving rare or noisy candidates.

\paragraph{Sliding-window retrieval.}

Given an input \texttt{0243} sequence, we first split the sequence into multiple code segments according to spaces. For each segment, sliding windows are applied to enumerate candidate sub-codes with lengths in $\{2,3,4\}$.

Empirically, Cantonese lyrics are dominated by bigram and trigram expressions, while many four-character candidates correspond to idiomatic or fixed expressions that are less compatible with free-form lyric generation. Therefore, the retrieval process mainly focuses on 2- and 3-character candidates.

The retrieval rules are summarized as follows:
\begin{itemize}
    \item Enumerate sub-codes with lengths 2, 3, and 4 using sliding windows.
    \item Prioritize 2- and 3-character codes.
    \item Retain at most one 4-character code for each segment.
    \item Retrieve the top-4 candidate words for each code according to frequency ranking.
    \item Retain at most 4 codes for each segment.
    \item Limit the total number of retrieved candidates to 16 for each sample.
    \item Remove duplicated candidates across different codes.
\end{itemize}

Single-character candidates are discarded by default because they are usually dominated by highly frequent function words and introduce significant retrieval noise.
\subsection{Retrieval Coverage and Representative Behavior}
\label{app:retrieval-cases}

On the training set, at least one retrieved candidate appears
verbatim in approximately 37.7\% of target lyrics. This indicates
that retrieval provides a sparse lexical prior rather than a complete
answer bank. Table~\ref{tab:retrieval-cases} illustrates two
representative behaviors.

\begin{table}[t]
\centering
\small
\setlength{\tabcolsep}{3pt}
\begin{tabular*}{\columnwidth}
{@{\extracolsep{\fill}}lcccc@{}}
\toprule
Case & Input plan & GT hit & Output use & Plan match \\
\midrule
Use     & \texttt{2433334}  & Yes & 2 & Exact \\
Non-use & \texttt{22342433} & Yes & 0 & Exact \\
\bottomrule
\end{tabular*}
\caption{Representative retrieval behavior. GT hit indicates that at
least one retrieved candidate occurs in the reference lyric. Output
use counts retrieved candidates appearing verbatim in the generated
lyric. Both outputs follow the input plan exactly, showing that
retrieved candidates are optional rather than copied answers.}
\label{tab:retrieval-cases}
\end{table}
\paragraph{Retrieval context formatting.}

The final retrieval context preserves the explicit alignment between \texttt{0243} sub-codes and candidate words:
\begin{equation}
R = \{(c_i,\mathcal{C}(c_i))\}_{i=1}^{M},
\end{equation}
where $c_i$ denotes the $i$-th retrieved \texttt{0243} sub-code and
$\mathcal{C}(c_i)$ denotes its associated candidate-word set.

The retrieval context is serialized into structured prompts using ``code : candidate words'' formatting:

\begin{quote}
\small
\ttfamily
44: candidate$_1$, candidate$_2$, candidate$_3$, candidate$_4$ \\
42: candidate$_1$, candidate$_2$, candidate$_3$, candidate$_4$ \\
20: candidate$_1$, candidate$_2$, candidate$_3$, candidate$_4$ \\
02: candidate$_1$, candidate$_2$, candidate$_3$, candidate$_4$
\ldots
\end{quote}

Instead of maximizing candidate recall, the proposed retrieval strategy aims to provide a compact set of high-frequency and distribution-consistent lexical priors while controlling prompt noise.

\section{Human Evaluation Details}
\label{app:user-study}

We conduct a human evaluation on 40 non-overlapping items from the
shared 500-item test subset. The study includes 12 volunteer
evaluators, with 6 male and 6 female participants. All participants
have a Cantonese-speaking background, can speak Cantonese and sing
Cantonese songs, and have undergraduate-level or higher education. All four systems are conditioned on the same reference
\texttt{0243} sequence; the vocal recording is presented only to
raters as a melody reference.

In the questionnaire, each sample is presented with its
\texttt{0243} sequence as the main melody--tone reference.
Arrow-based visual cues are provided to illustrate relative melodic
movement, and each sample is accompanied by a vocal audio reference.
Participants rate four anonymized system outputs along five dimensions:
tone--melody compatibility, rhythm, Cantonese naturalness, lyric
quality, and singability. They also select their preferred system
output when applicable.

All 9,600 dimension-level ratings are valid. For the preference
question, 413 responses are valid, 56 are marked as
\textit{uncertain}, and 11 are blank.
Table~\ref{tab:user-study-main} reports the descriptive mean scores
and valid preference counts.

\end{document}